\documentclass{article} 
\usepackage[preprint]{colm2026_conference}

\usepackage{microtype}
\usepackage{hyperref}
\usepackage{url}
\usepackage{booktabs}

\usepackage{graphicx}
\usepackage{subcaption}

\usepackage{lineno}

\usepackage{amsmath}
\usepackage{amssymb}
\usepackage{mathtools}
\usepackage{amsthm}

\usepackage{adjustbox}
\usepackage{tcolorbox}
\usepackage{wrapfig}

\usepackage{algorithm}
\usepackage{algorithmic}
\newcommand{\INPUT}{\item[\textbf{Input:}]}
\newcommand{\OUTPUT}{\item[\textbf{Output:}]}

\usepackage{tikz}

\usepackage{pifont}
\usepackage{xcolor}
\definecolor{mycoral}{HTML}{e07b54}
\definecolor{myblue}{HTML}{7fafd4}

\usepackage[capitalize,noabbrev]{cleveref}

\theoremstyle{plain}

\theoremstyle{definition}

\theoremstyle{remark}

\usepackage{relsize}
\usepackage{multirow}
\usepackage{enumitem}

\newcommand{\minisection}[1]{\vspace{.01in}\noindent{\textbf{#1}.}}

\definecolor{darkblue}{rgb}{0, 0, 0.5}
\hypersetup{colorlinks=true, citecolor=darkblue, linkcolor=darkblue, urlcolor=darkblue}

\title{SCOPE: Prompt Evolution for Enhancing Agent Effectiveness}

\author{
    Zehua Pei$^1$,  
    Hui-Ling Zhen$^2$, 
    Shixiong Kai$^2$, 
    Sinno Jialin Pan$^1$, \\
    \bf Yunhe Wang$^2$, 
    Mingxuan Yuan$^2$, 
    Bei Yu$^1$\\
    $^1$The Chinese University of Hong Kong \quad
    $^2$Huawei Technologies Co., Ltd
}

\begin{document}

\ifcolmsubmission
\linenumbers
\fi

\maketitle

\begin{abstract}
Large Language Model (LLM) agents are increasingly deployed in environments that generate massive, dynamic contexts.
However, a critical bottleneck remains: while agents have access to this context, their static prompts lack the mechanisms to manage it effectively, leading to recurring Corrective and Enhancement failures.
To address this capability gap, we introduce \textbf{S}elf-evolving \textbf{C}ontext \textbf{O}ptimization via \textbf{P}rompt \textbf{E}volution (\textbf{SCOPE}).
SCOPE frames context management as an \textit{online optimization} problem, synthesizing guidelines from execution traces to automatically evolve the agent's prompt.
We propose a Dual-Stream mechanism that routes guidelines between tactical memory (immediate error correction) and strategic memory, which is continuously refined through conflict resolution, subsumption pruning, and consolidation.
To maximize strategy coverage, Perspective-Driven Exploration evolves multiple parallel prompts guided by distinct optimization perspectives.
Experiments on the HLE benchmark show that SCOPE improves task success rates from 14.23\% to 38.64\% without human intervention.
We make our code publicly available at \url{https://github.com/JarvisPei/SCOPE}.

\end{abstract}

\section{Introduction}

The core functionality of a Large Language Model (LLM) agent is to perceive and react to context. 
Whether interpreting a user's ambiguous instruction, analyzing a screen full of code, or deciphering a traceback from a failed tool execution, the agent's success depends entirely on its ability to process this incoming information and decide on the next action.
As agents are tasked with increasingly complex problems, the volume and complexity of this context grow exponentially, such as in deep research~\citep{openai_ds,google_ds,wei2025browsecomp, zhang2025agentorchestra, team2025tongyi} and agentic coding~\citep{claude_code,jimenez2023swe}.
Consequently, the definition of a capable agent is shifting from one that simply knows facts to one that can effectively manage and navigate complex contexts.

\begin{figure*}[t]
\vspace{-1cm}
    \centering
    \includegraphics[width=0.88\linewidth]{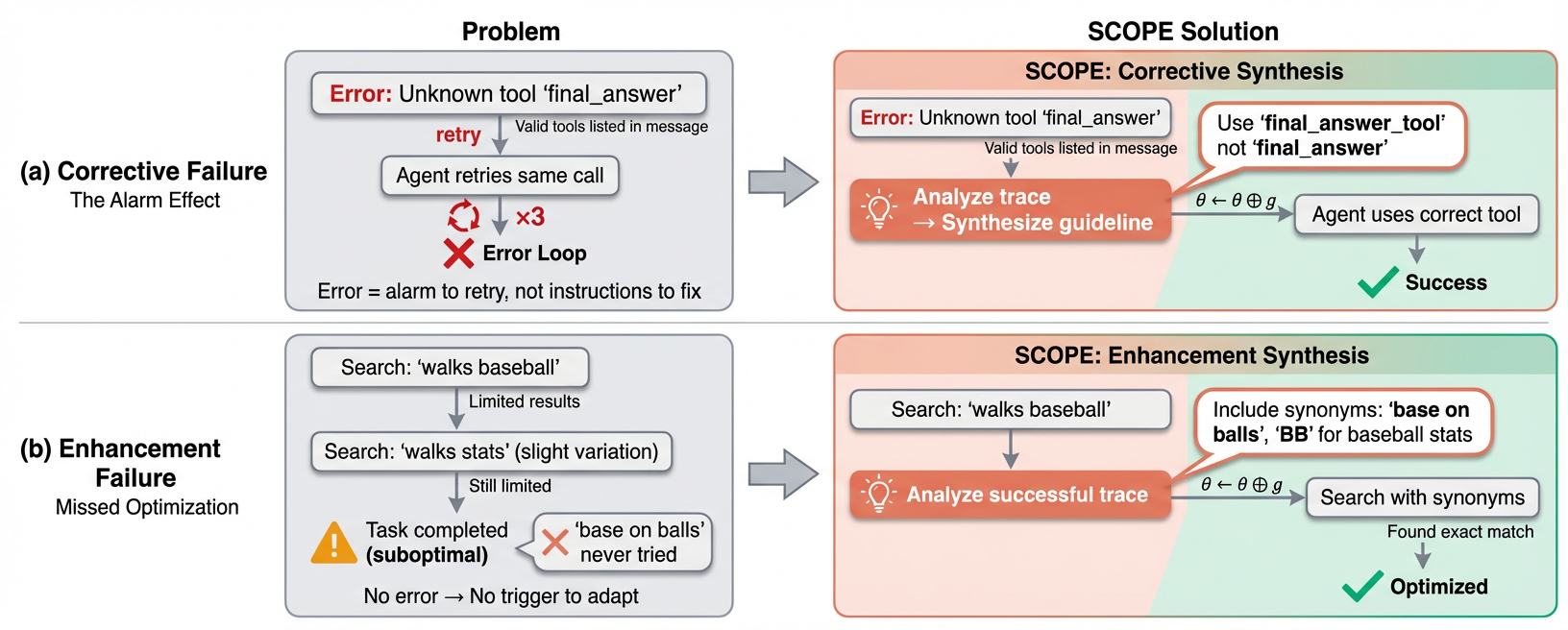}
    \caption{Two failure modes and how SCOPE addresses them. (a) \textbf{Corrective Failure}: The agent treats errors as generic alarms, entering error loops despite the error message containing the solution. SCOPE synthesizes a corrective guideline and integrates it into the prompt, enabling recovery. (b) \textbf{Enhancement Failure}: The agent persists with suboptimal strategies (\textit{e.g.}, single-term search) when no error is raised. SCOPE proactively analyzes successful traces to synthesize optimization guidelines.}
    \label{fig:failure_modes}
    \vspace{-3mm}
\end{figure*}

However, current research in agentic systems has predominantly focused on the availability of context rather than how agents manage it.
Innovations such as Retrieval-Augmented Generation (RAG)~\citep{lewis2020retrieval} or infinite context windows~\citep{team2024gemini} focus on feeding more data to the agent.
We argue that there is a fundamental gap between possessing context and dealing with context.
Even recent test-time learning methods like DC~\citep{suzgun2025dynamic} or ACE~\citep{zhang2025agentic}, which attempt to bridge this gap, operate at task-level granularity and cannot adapt during execution.
Without an evolved internal strategy to interpret these signals, an agent flooded with context is merely an agent flooded with noise.
Figure~\ref{fig:failure_modes} illustrates this gap: agents receive actionable information in their context but fail to leverage it effectively.

In this paper, we first substantiate this gap through an extensive analysis of over 1.5 million lines of execution logs from standard agents.
Our observations reveal that static prompts fail to adapt to dynamic context needs, leading to two distinct failure modes.
First, agents suffer from \textit{Corrective Failures}: when errors occur, agents treat error logs as generic alarms (signals to retry) rather than actionable feedback containing the solution.
In severe cases, agents fabricate data to proceed rather than managing uncertainty.
Second, agents exhibit \textit{Enhancement Failures}: even when operating without errors, agents miss opportunities to optimize, persisting with suboptimal strategies because their static prompts lack the mechanism to learn from execution patterns.

These failures are magnified in modern agentic systems, which are composed of specialized agents (\textit{e.g.}, coder, browser, analyzer) engaging in long, multi-turn interactions.
The complexity of these systems and the sheer volume of execution steps render static, one-size-fits-all prompts obsolete.
Yet this structure also creates an opportunity: since agents are invoked repeatedly, their prompts can evolve online.

To address this gap, we propose \textbf{S}elf-evolving \textbf{C}ontext \textbf{O}ptimization via \textbf{P}rompt \textbf{E}volution (\textbf{SCOPE}), a framework that transforms context management from a manual engineering task into an automatic optimization process.
SCOPE operates on the insight that the agent's own execution trace serves as the perfect training signal.
Through \textbf{Trace-Based Guideline Synthesis}, SCOPE analyzes these traces online to generate guidelines that teach the agent how to handle specific context patterns.
A \textbf{Dual-Stream Routing} mechanism routes guidelines to tactical (task-specific) or strategic (persistent) memory, where a \textbf{Memory Optimization} pipeline consolidates, prunes, and resolves conflicts to maintain a compact, high-quality guideline set.
Finally, \textbf{Perspective-Driven Exploration} evolves multiple parallel prompts guided by distinct perspectives, maximizing strategy coverage.
Empirically, 61\% of SCOPE's synthesized guidelines are proactive enhancements rather than reactive corrections, confirming that SCOPE functions primarily as an optimizer.

\vspace{-1mm}
Our contributions are as follows:
\vspace{-3mm}
\begin{itemize}
    \item We systematically study agent failures in modern agentic systems. We identify two failure modes, \textit{i.e.} Corrective Failures and Enhancement Failures, supported by analysis of over 1.5 million lines of execution logs.
    \vspace{-0.5mm}
    \item We propose SCOPE, a framework that synthesizes guidelines from execution traces, routes them via dual-stream routing with memory optimization, and explores diverse strategies through multiple perspectives.
    \vspace{-0.5mm}
    \item We demonstrate that SCOPE significantly outperforms static baselines, raising task success rates on the challenging HLE benchmark from 14.23\% to 38.64\% and on GAIA from 32.73\% to 56.97\%.
    \vspace{-4mm}
\end{itemize}


\section{Observations and Motivation}
\label{sec:observations}

We analyzed over 1.5 million lines of execution logs from baseline agents on the GAIA~\citep{mialon2023gaia} and DeepSearch~\citep{chen2025xbench} benchmarks.
Our findings reveal that agents often have the information needed to succeed, but their static prompts lack the mechanism to learn from execution feedback.
We categorize failures into two modes based on their trigger: \textit{Corrective Failures} occur when errors provide explicit signals that agents fail to act upon, while \textit{Enhancement Failures} occur when agents miss opportunities to optimize even in the absence of errors.
Figure~\ref{fig:failure_modes} illustrates both failure modes and how SCOPE addresses them through trace-based guideline synthesis.


\begin{figure*}[t]
\vspace{-1cm}
    \centering
    \includegraphics[width=0.85\linewidth]{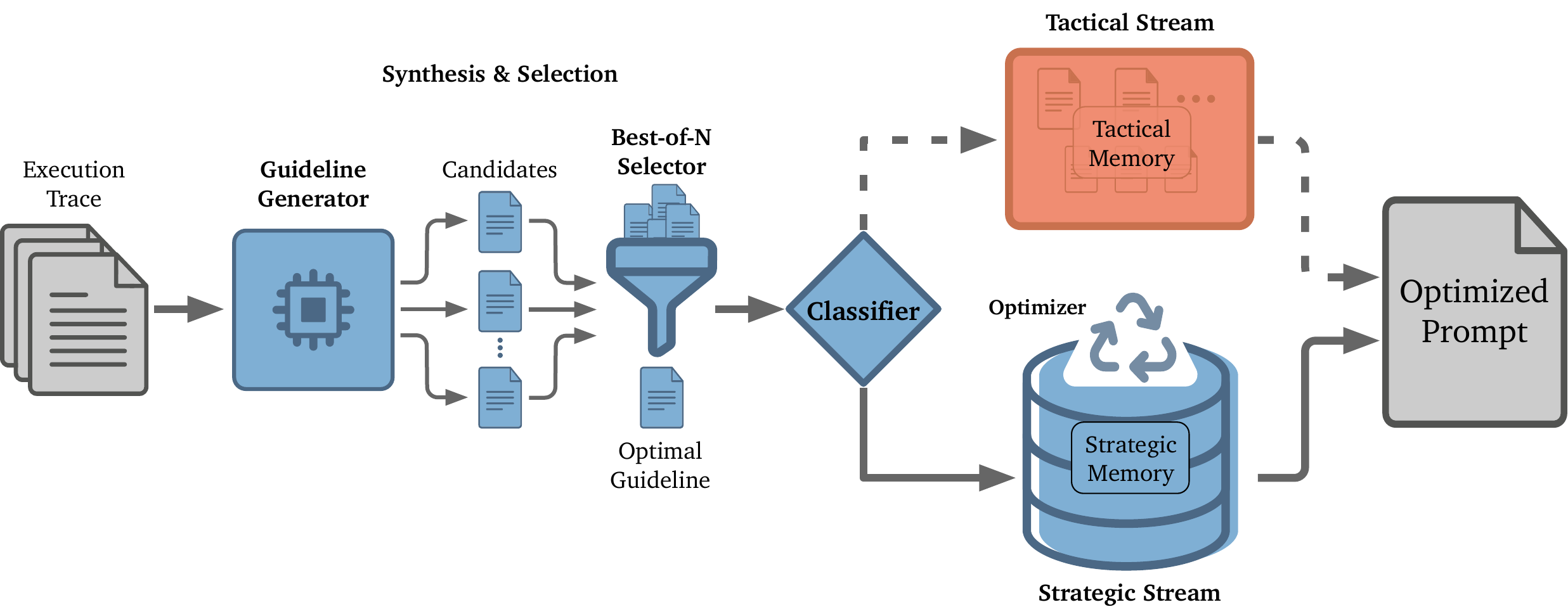}
    \caption{Overview of the SCOPE Framework. The \textbf{Generator} ($\pi_\phi$) synthesizes candidate guidelines from execution traces. A \textbf{Selector} ($\pi_\sigma$) chooses the best guideline, which is then routed by the \textbf{Classifier} ($\pi_\gamma$) to either \textbf{Tactical} (task-specific) or \textbf{Strategic} (persistent) memory. The \textbf{Optimizer} ($\pi_\omega$) consolidates the strategic memory.}
    \label{fig:method_overview}
    \vspace{-4mm}
\end{figure*}

\subsection{Failure Modes in Agentic Systems}
\label{sec:failure_modes}

\minisection{Corrective Failures}
\label{sec:corrective_failures}
When errors occur, execution traces contain explicit signals (error messages, stack traces, valid argument lists) that should guide correction. However, static agents treat these as generic ``alarms'' rather than actionable feedback. We observed over 70 instances where agents misused tools despite the error message explicitly listing valid usage, creating error loops where agents acknowledge failures but repeat the same action. In severe cases, agents respond to errors by fabricating data (\textit{e.g.}, ``Let's assume the file contains...''), \textit{i.e.} a catastrophic failure to manage uncertainty.

\minisection{Enhancement Failures}
\label{sec:enhancement_failures}
A subtler failure mode occurs when agents miss optimization opportunities even without explicit errors. Agents demonstrate rigid behavior: when search results are poor, context often implies synonyms would help (\textit{e.g.}, ``base on balls'' vs. ``walks''), yet agents persist with single keywords. We also observed silent quality issues, where agents misinterpreting domain concepts (\textit{e.g.}, confusing Eulerian and Hamiltonian paths) because static prompts provide no mechanism for additional verification.

\minisection{Complexity of Modern Agentic Systems}
\label{sec:complexity}
Modern benchmarks demand long, multi-role interactions: successful GAIA trajectories average 16.4 steps, with complex tasks exceeding 30 turns. Furthermore, systems comprise specialized sub-agents (\textit{e.g.}, browser, analyzer) with distinct failure patterns, \textit{e.g.} the Web Search agent drives suboptimal strategy errors while the Analyzer accounts for silent quality issues. This heterogeneity implies optimization must be agent-specific. However, this structure also creates an opportunity: agents are invoked repeatedly, forming a natural loop for online learning if lessons can be encoded into their prompts. A detailed taxonomy of failure patterns is provided in Appendix~\ref{app:detailed_observations}.


\subsection{From Limitations to Solution}
\label{sec:limitations_to_solution}

\textbf{Existing Methods and Their Limitations.}
Prior work on improving agent performance falls into three categories, each with fundamental limitations for modern agentic systems.

\textit{Memory-augmented methods} accumulate strategies into a playbook incorporated into the solver's prompt.
Examples include Dynamic Cheatsheet (DC)~\citep{suzgun2025dynamic}, Agentic Context Engineering (ACE)~\citep{zhang2025agentic}, and ReasoningBank~\citep{ouyang2025reasoningbank}.
These methods are designed for single-query question answering with custom solvers.
They operate at task-level granularity, \textit{i.e.} reflecting after each task attempt to update the playbook.
This means an agent cannot adapt during a difficult task; if it fails at step 5 of a 30-step task, it must complete (or abandon) the entire task before learning.
Furthermore, they maintain a one-for-all playbook that mixes strategies across scenarios and agent roles.

\textit{In-context correction methods} feed errors and self-generated feedback back into the agent's context to guide future steps.
Reflexion~\citep{shinn2023reflexion} is a classical example, with related approaches including Self-Refine~\citep{madaan2023self} and ReAct~\citep{yao2022react}.
These mechanisms can be adapted to agent systems.
However, feedback is appended to the conversation history rather than integrated into the agent's instructions.
The agent must infer corrections from this feedback, which often leads to error loops where the agent acknowledges mistakes but repeats them.
We characterize this as an alarm-based approach: the agent is alerted to errors but not taught how to fix them.

\textit{Offline prompt optimization methods} search for better prompts before deployment using training sets or evolutionary algorithms.
Classical examples include OPRO~\citep{yang2023large}, DSPy~\citep{khattab2023dspy}, and GEPA~\citep{agrawal2025gepa}.
While these methods create strong initial prompts, the resulting prompt remains static during inference.
If the agent encounters novel contexts or error patterns not represented in the training set, it cannot adapt.

Notably, all existing methods operate at task-level granularity: offline methods (OPRO, DSPy) optimize before deployment, while online methods (DC, ACE) update only between tasks.
None adapts the agent's prompt during execution, nor provides mechanisms for memory quality control or strategy diversification.

\textbf{Our Insight: Prompt Evolution for Agents.}
We propose a fundamentally different approach that leverages the structural properties of agentic systems.
Rather than accumulating external strategies (memory-augmented), relying on in-context inference (alarm-based), or optimizing offline (static), we treat the agent's prompt as an evolvable parameter that improves from execution traces online.
Since agents are invoked repeatedly, their prompts can evolve throughout execution, where each invocation is an opportunity to apply learned guidelines.
Guidelines synthesized from traces are integrated directly into each agent's prompt, teaching the agent how to handle specific situations rather than hoping it infers corrections.
This approach naturally enables \textbf{step-level adaptation}, where the prompt is updated during execution to allow recovery from mid-task failures, and \textbf{per-agent optimization}, where each agent role evolves its own prompt based on role-specific patterns. Empirical validation of these design choices is provided in Appendix~\ref{app:further_analysis}.

\textbf{Problem Formulation.}
\label{sec:formulation}
We formalize this as a prompt optimization problem.
Consider an agent whose behavior is governed by a prompt $\theta$.
The execution generates a trace $\tau_t$ containing actions and observations.
We treat $\tau_t$ as a learning signal, \textit{i.e.} similar to how gradients guide optimization, where errors and suboptimal behaviors indicate how $\theta$ should be updated.
Since the prompt space is discrete, we cannot compute gradients directly.
Instead, we synthesize a natural language \textit{guideline} $g_t$ from the trace $\tau_t$:
\begin{equation}
    \theta_{t+1} = \theta_t \oplus g_t
\end{equation}
where $\theta_t$ is the current prompt, $g_t$ is the synthesized guideline, and $\oplus$ denotes integration.
This formulation shifts from static prompt engineering to self-evolving prompts that enhance agent effectiveness.

\section{Methodology}
\label{sec:method}

We introduce \textbf{S}elf-evolving \textbf{C}ontext \textbf{O}ptimization via \textbf{P}rompt \textbf{E}volution (\textbf{SCOPE}), a framework that implements the prompt optimization formulated in Section~\ref{sec:observations}.
SCOPE evolves the prompt $\theta$ in response to the execution trace $\tau$, synthesizing guidelines that improve agent effectiveness.
Our framework, illustrated in Figure~\ref{fig:method_overview}, consists of four components: (1) \textbf{Guideline Synthesis}, (2) \textbf{Dual-Stream Routing}, (3) \textbf{Memory Optimization}, and (4) \textbf{Perspective-Driven Exploration}. The complete procedure is outlined in Algorithm~\ref{alg:scope}.

\subsection{Guideline Synthesis}
\label{sec:synthesis}

The core of SCOPE is synthesizing guidelines from the execution trace.
We use a generator $\pi_\phi$ to produce candidate guidelines, followed by a selector $\pi_\sigma$ to choose the best one.

\textbf{Corrective Synthesis.}
When the agent encounters an error, the generator analyzes the trace using corrective rubrics $\mathcal{I}_{\text{corr}}$ (see Appendix~\ref{app:corr_rubrics}) to synthesize corrective guidelines:
\begin{equation}
    g = \pi_\phi(\tau_t, \theta_t, \mathcal{I}_{\text{corr}}).
    \label{eq:synthesis_corr}
\end{equation}
Conditioning on $\theta_t$ avoids redundant updates (\textit{e.g.}, re-learning an existing guideline).

\begin{wrapfigure}{r}{0.5\textwidth}
\vspace{-0.5cm}
\begin{minipage}{0.5\textwidth}
\begin{algorithm}[H]
\small
   \caption{SCOPE: Self-Evolving Prompt Optimization}
   \label{alg:scope}
\begin{algorithmic}
   \INPUT Task, Base Prompt $\theta_{\text{base}}$, Strategic Memory $\mathcal{M}_{\text{strat}}$, Rubrics $\mathcal{I}$
   \OUTPUT Completed Task, Updated $\mathcal{M}_{\text{strat}}$
   \STATE Init $\mathcal{M}_{\text{tact}} \!\leftarrow\! \emptyset$, $\theta_t \!\leftarrow\! \theta_{\text{base}} \!\oplus\! \mathcal{M}_{\text{strat}}$
   \WHILE{Task not completed}
       \STATE Execute with $\theta_t$, update $\tau_t$
       \IF{Error or Sub-task completion}
           \item[] \textit{// 1. Guideline Synthesis}
           \STATE $G \!\leftarrow\! \pi_\phi(\tau_t, \theta_t, \mathcal{I}_{\text{corr/enh}})$
           \STATE $g^* \!\leftarrow\! \pi_\sigma(G, \theta_t, \mathcal{I}_{\text{sel}})$
           \item[] \textit{// 2. Classification \& Routing}
           \STATE $(c, \text{conf}) \!\leftarrow\! \pi_\gamma(g^*, \theta_t, \mathcal{I}_{\text{cls}})$
           \IF{$c\!=\!\text{Tact.}$ OR $\text{conf}\!<\!c_{\text{thresh}}$}
               \STATE $\mathcal{M}_{\text{tact}} \!\leftarrow\! \mathcal{M}_{\text{tact}} \cup \{g^*\}$
           \ELSE
               \STATE $\mathcal{M}_{\text{strat}} \!\leftarrow\! \pi_\omega(\mathcal{M}_{\text{strat}} \!\cup\! \{g^*\})$
           \ENDIF
           \item[] \textit{// 3. Update Prompt}
           \STATE $\theta_{t+1} \!\leftarrow\! \theta_{\text{base}} \!\oplus\! \mathcal{M}_{\text{strat}} \!\oplus\! \mathcal{M}_{\text{tact}}$
       \ENDIF
   \ENDWHILE
\end{algorithmic}
\end{algorithm}
\end{minipage}
\vspace{-0.2cm}
\end{wrapfigure}

\textbf{Enhancement Synthesis.}
When execution succeeds but appears suboptimal, the generator uses enhancement rubrics $\mathcal{I}_{\text{enh}}$ (see Appendix~\ref{app:enh_rubrics}) to identify inefficiencies and synthesize enhancement guidelines:
\begin{equation}
    g = \pi_\phi(\tau_t, \theta_t, \mathcal{I}_{\text{enh}}).
    \label{eq:synthesis_enh}
\end{equation}

\textbf{Best-of-N Selection.}
To reduce variance, the generator produces $N$ candidates $G = \{g_1, \ldots, g_N\}$.
A selector $\pi_\sigma$ then chooses the best guideline according to selection rubrics $\mathcal{I}_{\text{sel}}$ (see Appendix~\ref{app:sel_rubrics}):
\begin{equation}
    g^* = \pi_\sigma(G, \theta_t, \mathcal{I}_{\text{sel}}).
    \label{eq:bon}
\end{equation}

\subsection{Dual-Stream Routing}
\label{sec:dual_stream}

After synthesis, each guideline $g^*$ is routed to one of two memory streams based on its scope.
We maintain a strategic memory $\mathcal{M}_{\text{strat}} = \{g_1, g_2, \ldots\}$ for long-term guidelines and a tactical memory $\mathcal{M}_{\text{tact}}$ for task-specific guidelines.

A classifier $\pi_\gamma$ evaluates the guideline's generality and assigns a confidence score according to classification rubrics $\mathcal{I}_{\text{cls}}$ (see Appendix~\ref{app:cls_rubrics}):
\begin{equation}
    (c, \text{conf}) = \pi_\gamma(g^*, \theta_t, \mathcal{I}_{\text{cls}}),
    \label{eq:classify}
\end{equation}
where $c \in \{\text{Tactical}, \text{Strategic}\}$ and $\text{conf} \in [0,1]$.

\textbf{Tactical Stream.}
Guidelines classified as tactical, or strategic with low confidence ($\text{conf} < c_{\text{thresh}}$), are added to the tactical memory:
\begin{equation}
    \mathcal{M}_{\text{tact}} \leftarrow \mathcal{M}_{\text{tact}} \cup \{g^*\}.
\end{equation}
These guidelines are valid only for the current task.

\textbf{Strategic Stream.}
High-confidence guidelines ($\text{conf} \ge c_{\text{thresh}}$) identifying general principles are added to the strategic memory:
\begin{equation}
    \mathcal{M}_{\text{strat}} \leftarrow \mathcal{M}_{\text{strat}} \cup \{g^*\}.
\end{equation}
These guidelines persist across tasks.

The prompt evolves by combining both memories:
\begin{equation}
    \theta_{t+1} = \theta_{\text{base}} \oplus \mathcal{M}_{\text{strat}} \oplus \mathcal{M}_{\text{tact}},
    \label{eq:prompt_update}
\end{equation}
where $\theta_{\text{base}}$ is the initial prompt before any guidelines are added.

\subsection{Memory Optimization}
\label{sec:quality}

Unconstrained growth of $\mathcal{M}_{\text{strat}}$ can dilute the agent's attention with redundant or conflicting guidelines.
We apply an optimizer $\pi_\omega$, which is a multi-step pipeline (see Appendix~\ref{app:opt_rubrics} for complete rubrics), to consolidate the memory:
\begin{equation}
    \mathcal{M}_{\text{strat}} \leftarrow \pi_\omega(\mathcal{M}_{\text{strat}} \cup \{g^*\}).
\end{equation}
The pipeline consists of three steps:
(1) \textbf{Conflict Resolution}: merging contradictory guidelines.
(2) \textbf{Subsumption Pruning}: removing specific guidelines covered by general ones.
(3) \textbf{Consolidation}: merging similar guidelines into comprehensive ones.
See Appendix~\ref{app:memory_example} for a concrete example.

\subsection{Perspective-Driven Exploration}
\label{sec:global_bon}

A single evolved prompt may converge to a strategy that works well on some tasks but poorly on others.
To increase coverage, we initialize $K$ parallel streams with distinct ``Perspectives'' (\textit{e.g.}, Efficiency vs. Thoroughness), each evolving its own prompt $\theta^k$.
At test time, we select the best result:
\begin{equation}
    \text{Success} = \max_{k \in \{1..K\}} \text{Eval}(\theta^k, \text{task}),
\end{equation}
where $\text{Eval}(\theta^k, \text{task})$ returns whether the agent with prompt $\theta^k$ successfully completes the task.
This allows the system to leverage diverse strategies for different problem types. 


\section{Experiments}
\label{sec:exp}

\subsection{Experimental Setup}
\label{sec:exp_setup}

\minisection{Baseline Agent System}
We implement a hierarchical agent system with a planning agent that delegates to specialized subordinate agents (web search, analyzer, browser), each equipped with domain-specific tools. We use Gemini-2.5-Pro for the web search and analyzer agents, and GPT-4.1 for the planning and browser agents. Details are in Appendix~\ref{app:agent_system}.

\minisection{Benchmarks}
We evaluate on three benchmarks representative of the Deep Research paradigm~\citep{openai_ds,google_ds}, \textit{i.e.} autonomous agents that perform multi-step web research, tool use, and information synthesis: \textbf{HLE}~\citep{phan2025humanity} (2,500 expert-level questions), \textbf{GAIA}~\citep{mialon2023gaia} (165 multi-tool reasoning tasks), and \textbf{DeepSearch}~\citep{chen2025xbench} (multi-hop retrieval and cross-document synthesis). Details are in Appendix~\ref{app:benchmarks}.

\minisection{SCOPE Configuration}
We configure SCOPE with $N=2$ candidates for Best-of-N synthesis and $K=2$ parallel streams (Efficiency and Thoroughness) for global exploration. The strategic memory is capped at 10 guidelines per domain, triggering consolidation when exceeded, with a confidence threshold of 0.85 for promotion. All SCOPE components (Generator, Selector, Classifier, Optimizer) utilize GPT-4.1. Additionally, we implement Agent-Specific Optimization, step-level updates, and place Synthesized guidelines in the system prompt (see Section~\ref{sec:placement} for ablation). Detailed hyperparameter settings are in Appendix~\ref{app:hyperparams}.

\minisection{Prompt Optimization Baselines}
We compare SCOPE against two recent prompt optimization methods:
\textbf{(1) Dynamic Cheatsheet (DC)}~\citep{suzgun2025dynamic}: A test-time learning framework that endows LLMs with persistent, evolving memory. 
We implement the DC-Cu variant, which cumulatively updates memory after processing each input.
\textbf{(2) Agentic Context Engineering (ACE)}~\citep{zhang2025agentic}: A playbook-based learning approach that maintains bullet-point strategies across predefined categories and updates them via a reflector-curator loop. We implement ACE adapting it to our baseline agent system with step-level granularity for fair comparison.

\subsection{Overall Performance and Ablation}
\label{sec:results_ablation}

Table~\ref{tab:main_results} and Table~\ref{tab:ablation} present the main experimental results and component analysis. All methods are evaluated with two independent runs per task; a task is considered solved if either run succeeds (Pass@2). SCOPE establishes a new state-of-the-art across all benchmarks, significantly outperforming both the static baseline and recent optimization methods like DC~\citep{suzgun2025dynamic} and ACE~\citep{zhang2025agentic}. On the expert-level HLE benchmark, SCOPE more than doubles the baseline performance (38.64\% vs. 14.23\%).

The ablation study (Table~\ref{tab:ablation}) confirms that every component contributes to this success. While the basic Guideline Generator provides the initial boost (+4.85\%), the largest gain comes from our Perspective-Driven Exploration ($K=2$), which adds 10.91\%. This underscores that a single context management strategy is insufficient; maintaining diverse optimization streams (Efficiency and Thoroughness) is critical for robust performance across heterogeneous tasks.

\begin{table}[tb!]
\vspace{-1cm}
\centering
\begin{minipage}[t]{0.48\linewidth}
    \centering
    \caption{Performance comparison on three challenging agent benchmarks.}
    \label{tab:main_results}
    \resizebox{\linewidth}{!}{
    \begin{tabular}{lccc}
        \toprule
        \textbf{Method} & \textbf{HLE} & \textbf{GAIA} & \textbf{DeepSearch} \\
        \midrule
        \multicolumn{4}{l}{\textit{Base Models}} \\
        \midrule
        GPT-4.1 & 6.00 & 8.48 & 4.00 \\
        Gemini-2.5-Pro & 18.76 & 26.67 & 19.00 \\
        \midrule
        \multicolumn{4}{l}{\textit{Agent Systems}} \\
        \midrule
        Baseline Agent & 14.23 & 32.73 & 14.00 \\
        DC~\citep{suzgun2025dynamic} & 18.44 & 35.76 & 21.00 \\
        ACE~\citep{zhang2025agentic} & 23.72 & 38.18 & 23.00 \\
        \textbf{SCOPE} & \textbf{38.64} & \textbf{56.97} & \textbf{32.00} \\
        \bottomrule
    \end{tabular}
    }
\end{minipage}
\hfill
\begin{minipage}[t]{0.48\linewidth}
    \centering
    \caption{Ablation study on GAIA. Each row adds a component cumulatively.}
    \label{tab:ablation}
    \resizebox{\linewidth}{!}{
    \begin{tabular}{lcc}
        \toprule
        \textbf{Configuration} & \textbf{Acc. (\%)} & \textbf{$\Delta$} \\
        \midrule
        Baseline Agent (Static) & 32.73 & - \\
        + Guideline Generator & 37.58 & \textcolor{mycoral}{+4.85} \\
        + Dual-Stream Routing & 41.21 & \textcolor{mycoral}{+3.63} \\
        + Best-of-N Selection & 44.24 & \textcolor{mycoral}{+3.03} \\
        + Memory Optimization & 46.06 & \textcolor{mycoral}{+1.82} \\
        + Perspective Exploration& \textbf{56.97} & \textcolor{mycoral}{+10.91} \\
        \bottomrule
    \end{tabular}
    }
\end{minipage}
\vspace{-0.5cm}
\end{table}


\subsection{Guideline Placement: System Prompt vs. User Prompt}
\label{sec:placement}

\begin{wraptable}{r}{0.52\textwidth}
\vspace{-1.2em}
    \centering
    \caption{Impact of guideline placement on GAIA. All placements reduce errors vs. baseline (15--42\%), but system prompt achieves the best accuracy.}
    \label{tab:placement}
    \resizebox{\linewidth}{!}{
    \begin{tabular}{lcccc}
        \toprule
        \textbf{Placement} & \textbf{Acc. (\%)} & \textbf{Errors} & \textbf{Max Steps} & \textbf{Steps} \\
        \midrule
        Baseline & 32.73 & 1,714 & 255 & 9,824 \\
        \midrule
        (1) System Prompt & \textbf{46.06} & 1,461 \textcolor{mycoral}{\scriptsize(-15\%)} & 227 & 7,430 \\
        (2) User Prompt & 41.21 & 1,000 \textcolor{mycoral}{\scriptsize(-42\%)} & 130 & 7,408 \\
        (3) Split  & 35.76 & 1,204 \textcolor{mycoral}{\scriptsize(-30\%)} & 141 & 7,319 \\
        (4) Hybrid  & 43.64 & 1,109 \textcolor{mycoral}{\scriptsize(-35\%)} & 139 & 6,620 \\
        \bottomrule
    \end{tabular}
    }
\vspace{-1em}
\end{wraptable}

Table~\ref{tab:placement} compares four placement strategies on GAIA: (1) all in system prompt, (2) all in user prompt, (3) split (strategic in system, tactical in user), and (4) hybrid (saved in system, online in user). A counter-intuitive finding emerges: system prompt placement achieves the highest accuracy (46.06\%) despite more tasks hitting the step limit than user prompt (227 vs. 130). We hypothesize that system prompt guidelines act as \textit{implicit background guidance}, allowing the agent to internalize guidelines and explore more solution paths. Conversely, user prompt placement leads to \textit{over-compliance}---the agent follows instructions too strictly, terminating early rather than continuing exploration, which results in fewer errors but lower task success. The split strategy performs worst (35.76\%), as distributing guidelines across both prompts creates conflicting priorities. Detailed analysis: Appendix~\ref{app:placement_analysis}.

\subsection{Model Choice and Computational Overhead}
\label{sec:model_ablation}

A practical deployment question is which model to use for SCOPE's components (Generator, Classifier, Optimizer). Since our baseline agent uses both GPT-4.1 and Gemini-2.5-Pro for different roles, we evaluate three configurations: (1) all components use GPT-4.1, (2) all use Gemini-2.5-Pro, and (3) each meta-agent uses the same model as the base agent it optimizes.

Table~\ref{tab:model_ablation} reveals a surprising finding: all three configurations achieve nearly identical performance (within 1.2\%), despite Gemini generating 46\% more guidelines than GPT-4.1. This suggests that guideline quality matters more than quantity, \textit{i.e.} SCOPE's selection and optimization mechanisms effectively filter for useful guidelines regardless of how many candidates are generated. This robustness to model choice simplifies deployment, allowing practitioners to select meta-agent models based on cost or latency rather than accuracy concerns. Further analysis of guideline distribution patterns is provided in Appendix~\ref{app:model_ablation}.

Table~\ref{tab:cost} presents a detailed overhead analysis on 100 HLE samples. SCOPE adds only 4.2\% more API calls compared to the baseline. While prompt tokens increase 152\% due to accumulated guidelines, completion tokens \textit{decrease} by 38.6\%, indicating fewer wasted reasoning turns. Notably, SCOPE's synthesis runs asynchronously, adding zero wall-clock latency per step, unlike the reflection baseline, which adds $\sim$10s per step and uses 108\% more API calls yet scores only 16.00 vs.\ SCOPE's 36.00, confirming that gains stem from \textit{what} is learned, not how much is computed.

\subsection{SCOPE Enhances Robustness in Long-Horizon Domains}
\label{sec:robustness}

\begin{figure*}[tb!]
    \centering
    \vspace{-2mm}
    \includegraphics[width=0.95\linewidth]{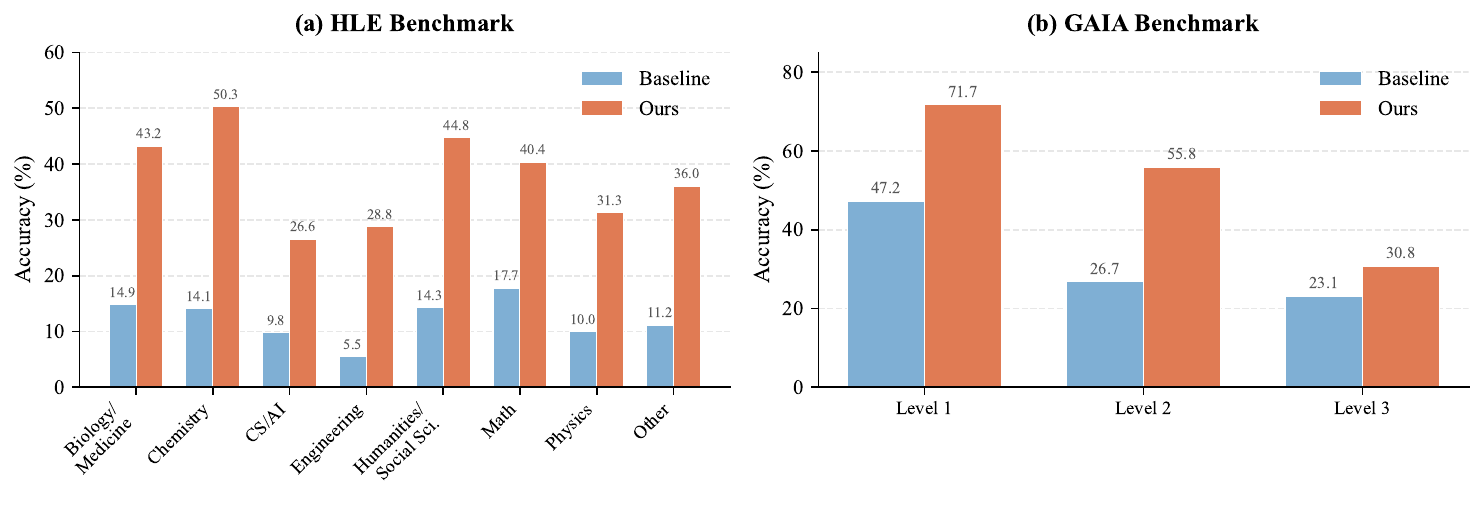}
    \vspace{-3mm}
    \caption{Performance breakdown by subcategory on HLE (a) and GAIA (b). SCOPE (\textcolor{mycoral}{coral}) consistently outperforms the baseline (\textcolor{myblue}{blue}) across all domains. 
    }
    \label{fig:subcategory_results}
    \vspace{-0.5cm}
\end{figure*}

Figure~\ref{fig:subcategory_results} breaks down performance by subcategory. On HLE, the largest gains are in knowledge-intensive domains: Biology/Medicine (14.9\% $\rightarrow$ 43.2\%) and Chemistry (14.1\% $\rightarrow$ 50.3\%), where specialized tool errors are common and SCOPE's domain-specific guidelines enable recovery. On GAIA Level 3 tasks, \textit{i.e.} the longest trajectories with most noise, SCOPE achieves 30.8\% vs 23.1\%, confirming that dynamic prompts prevent error propagation and maintain coherence in long-horizon scenarios.

\begin{table}[tb!]
\vspace{-1cm}
\centering
\begin{minipage}[t]{0.40\linewidth}
    \centering
    \caption{Model choice for SCOPE's meta-agents on GAIA.}
    \label{tab:model_ablation}
    \resizebox{\linewidth}{!}{
    \begin{tabular}{lccc}
        \toprule
        \textbf{Meta-Agent Model} & \textbf{Acc. (\%)} & \textbf{Guidelines} & \textbf{Avg. Len.} \\
        \midrule
        All GPT-4.1 & 46.06 & 111 & 380 chars \\
        All Gemini-2.5-Pro & \textbf{46.67} & 163 & 426 chars \\
        Same as Base Agent & 45.45 & 108 & 397 chars \\
        \bottomrule
    \end{tabular}
    }
\end{minipage}
\hfill
\begin{minipage}[t]{0.57\linewidth}
    \centering
    \caption{Computational overhead on 100 HLE samples.}
    \label{tab:cost}
    \resizebox{\linewidth}{!}{
    \begin{tabular}{lccccc}
        \toprule
        \textbf{Method} & \textbf{HLE Acc.} & \textbf{API Calls} & \textbf{Prompt Tok.} & \textbf{Compl. Tok.} & \textbf{Synth. Lat.} \\
        \midrule
        Baseline & 15.00 & 1,119 & 4.9M & 469K & --- \\
        Baseline+Refl. & 16.00 & 2,331 {\scriptsize(+108\%)} & 14.9M {\scriptsize(+203\%)} & 432K {\scriptsize($-$8\%)} & $\sim$10s/step \\
        \textbf{SCOPE} & \textbf{36.00} & 1,166 {\scriptsize(+4.2\%)} & 12.4M {\scriptsize(+152\%)} & 288K {\scriptsize($-$38.6\%)} & \textbf{+0s (async)} \\
        \bottomrule
    \end{tabular}
    }
\end{minipage}
\end{table}

Additional experiments confirm that SCOPE generalizes to single-agent coding (SWE-bench Verified~\citep{jimenez2023swe}, +8.2 points; Appendix~\ref{app:swe_bench}) and scales with stronger frontier models (GPT-5.4, +17.5 points; Appendix~\ref{app:gpt54}).
\section{Analysis}
\label{sec:analysis}

\subsection{Mitigating Agentic Fragility via Dual-Mode Synthesis}
\label{sec:corrective_enhancement}

A counter-intuitive finding in Table~\ref{tab:main_results} is that the Baseline Agent underperforms raw Gemini-2.5-Pro on HLE (14.23\% vs. 18.76\%) and DeepSearch (14.00\% vs. 19.00\%). While agentic wrappers enable tool use, they introduce structural fragility: repeated tool errors or infinite loops can cause a capable model to fail tasks it could answer zero-shot. Table~\ref{tab:placement} quantifies this on GAIA: the baseline accumulates 1,714 errors across 165 tasks.

SCOPE addresses this fragility through two complementary synthesis modes: \textbf{Corrective} (reactive error correction) and \textbf{Enhancement} (proactive optimization from successful patterns). For instance, a \texttt{NameError} triggers a corrective guideline (``Define all variables in code snippets''), while a successful search triggers an enhancement guideline (``Try search term variants''). The corrective mode enables real-time ``debugging'', where synthesizing guidelines from failure traces reduces errors by 15--42\% depending on placement strategy (Table~\ref{tab:placement}), explaining why SCOPE exceeds base model capabilities.

Crucially, Table~\ref{tab:trigger_distribution} shows that enhancement guidelines constitute 61\% of all synthesized guidelines, demonstrating that SCOPE is predominantly an optimizer rather than merely an error debugger. This prevalence of proactive guidelines explains why SCOPE continues to improve even when error rates are low, \textit{i.e.} it codifies successful patterns to prevent \textit{potential} failures before they occur.

\subsection{Guideline Compliance: Do Agents Follow Synthesized Guidelines?}
\label{sec:compliance}

A critical question is whether synthesized guidelines actually influence agent behavior. Figure~\ref{fig:compliance_trace} (Appendix) illustrates what we term language adoption: after SCOPE synthesizes a guideline recommending ``plausible label synonyms and phrase variants'', the agent's subsequent output incorporates this exact phrasing verbatim. This direct linguistic transfer provides strong evidence that guidelines are actively integrated into the agent's decision-making, not merely stored.
Beyond linguistic adoption, we observe immediate behavioral changes (typically within seconds) following guideline synthesis, demonstrating real-time adaptation within a single episode. 

\begin{table}[tb!]
\vspace{-1cm}
\centering
\begin{minipage}[t]{0.48\linewidth}
    \centering
    \caption{Distribution of synthesized guidelines by mode and domain on GAIA. Enhancement guidelines (61\%) dominate.}
    \label{tab:trigger_distribution}
    \resizebox{\linewidth}{!}{
    \begin{tabular}{lccc}
        \toprule
        \textbf{Domain} & \textbf{Corrective} & \textbf{Enhancement} & \textbf{Total} \\
        \midrule
        Analysis Methodology & 9 & 22 & 31 \\
        Tool Usage & 15 & 13 & 28 \\
        Data Validation & 8 & 15 & 23 \\
        Efficiency & 2 & 16 & 18 \\
        Error Handling & 9 & 2 & 11 \\
        \midrule
        \textbf{Total} & \textbf{43 (39\%)} & \textbf{68 (61\%)} & \textbf{111} \\
        \bottomrule
    \end{tabular}
    }
\end{minipage}
\hfill
\begin{minipage}[t]{0.50\linewidth}
    \centering
    \caption{Performance of individual perspectives vs. Global Ensemble on GAIA. Low overlap (33.94\%) validates diverse optimization.}
    \label{tab:perspective_diversity}
    \resizebox{\linewidth}{!}{
    \begin{tabular}{lcccc}
        \toprule
        \textbf{Perspective} & \textbf{Lv.1} & \textbf{Lv.2} & \textbf{Lv.3} & \textbf{Total} \\
        \midrule
        Efficiency & 60.38 & 40.70 & 26.92 & 44.85 \\
        Thoroughness & 60.38 & 47.67 & 11.54 & 46.06 \\
        \midrule
        \textbf{Intersection} & 49.06 & 32.56 & 7.69 & 33.94 \\
        \textbf{Global Ensemble} & \textbf{71.70} & \textbf{55.81} & \textbf{30.77} & \textbf{56.97} \\
        \bottomrule
    \end{tabular}
    }
\end{minipage}
\vspace{-0.5cm}
\end{table}

\subsection{Perspective Divergence in Guideline Synthesis}
\label{sec:qualitative_analysis}

Table~\ref{tab:perspective_diversity} validates our Perspective-Driven Exploration design by analyzing the two perspectives. While both streams achieve similar total accuracy (44.85\% vs 46.06\%), their intersection is only 33.94\%, \textit{i.e.} approximately 23\% of solved problems are unique to one perspective. Notably, Efficiency outperforms on Level 3 tasks (26.92\% vs 11.54\%), suggesting concise context management is more effective for complex, long-horizon tasks, while Thoroughness excels at Level 2. Global Ensemble captures the union of strengths.

This quantitative divergence stems from fundamentally different guideline strategies. As illustrated in Figure~\ref{fig:qualitative_example} (Appendix), when the browser agent encounters access denials (HTTP 403), the Efficiency stream learns to ``fail-over fast'' by escalating to the web search agent, while the Thoroughness stream synthesizes guidelines to ``find alternate sources'' (e.g., Archive.org). This duality allows SCOPE to handle both time-constrained and deep retrieval tasks. A detailed qualitative comparison is provided in Appendix~\ref{app:perspective_qualitative}.

\section{Related Work}


\minisection{Context Augmentation and Compression}
The primary challenge in agentic workflows is handling extensive information.
Traditional approaches address this via retrieval and compression.
Retrieval-Augmented Generation (RAG)~\citep{lewis2020retrieval} fetches relevant context, while methods like LLMLingua~\citep{jiang2023llmlingua, sun2025scaling} optimize efficiency via compression.
They focus on what context to provide, whereas SCOPE focuses on how the agent processes context by evolving prompt.

\minisection{Prompt Optimization}
A significant body of work focuses on prompt engineering.
Offline methods like OPRO~\citep{yang2023large}, DSPy~\citep{khattab2023dspy}, and GEPA~\citep{agrawal2025gepa} search for optimal prompts before deployment using training sets.
While these create strong initializations, the resulting prompt remains static during inference.
SCOPE complements these by enabling online adaptation: the prompt evolves during execution based on step-level feedback.

\minisection{Agent Memory and Learning}
Recent work gives agents persistent memory to accumulate experience.
Memory-augmented methods like DC~\citep{suzgun2025dynamic}, ACE~\citep{zhang2025agentic}, and ReasoningBank~\citep{ouyang2025reasoningbank} build external strategy libraries.
In-context correction methods like Reflexion~\citep{shinn2023reflexion} and Self-Refine~\citep{madaan2023self} feed feedback into the context.
As discussed in Section~\ref{sec:limitations_to_solution}, they face limitations in agentic settings: task-level granularity, one-for-all memory, and alarm-based correction.
SCOPE addresses these with step-level adaptation, per-agent optimization, and prompt evolution.

\section{Conclusion}

We introduced \textbf{SCOPE}, a framework that enables prompt evolution for enhancing agent effectiveness.
By treating execution traces as learning signals, SCOPE synthesizes guidelines that are integrated directly into agent prompts, enabling step-level adaptation and per-agent optimization.
Experiments on HLE and GAIA demonstrate that SCOPE significantly outperforms static baselines and existing methods, more than doubling success rates in expert-level domains.
Our findings suggest a new direction: rather than engineering static prompts, we should enable agents to evolve their own prompts online.

\bibliography{ref/main, ref/example_paper}
\bibliographystyle{colm2026_conference}

\clearpage
\appendix
\section*{Appendix Overview}

This appendix provides supplementary materials organized as follows:

\begin{itemize}[leftmargin=*,itemsep=2pt]
    \item \textbf{Appendix~\ref{app:detailed_observations}: Detailed Analysis of Baseline Failure Modes} --- Comprehensive taxonomy and examples of agent failures from over 1.5 million lines of execution logs.
    \item \textbf{Appendix~\ref{app:agent_system}: Baseline Agent System Details} --- Architecture and configuration of our hierarchical multi-agent system.
    \item \textbf{Appendix~\ref{app:benchmarks}: Benchmark Details} --- Description of HLE, GAIA, and DeepSearch evaluation benchmarks.
    \item \textbf{Appendix~\ref{app:hyperparams}: Hyperparameter Settings} --- Complete hyperparameters, domain categories, optimization pipeline details, and a concrete memory optimization example.
    \item \textbf{Appendix~\ref{app:global_diversity}: Perspective-Driven Exploration Details} --- Implementation of Efficiency and Thoroughness optimization streams.
    \item \textbf{Appendix~\ref{app:prompts}: Complete System Prompts} --- Full rubrics for all meta-agents (Generator, Selector, Classifier, Optimizer).
    \item \textbf{Appendix~\ref{app:further_analysis}: Per-Agent Specialization and Step-Level Updates} --- Empirical validation of per-agent and step-level design choices.
    \item \textbf{Appendix~\ref{app:placement_analysis}: Guideline Placement Analysis} --- Deep dive into system vs. user prompt placement effects.
    \item \textbf{Appendix~\ref{app:perspective_qualitative}: Qualitative Comparison of Perspective Streams} --- Detailed strategy patterns for Efficiency vs. Thoroughness streams.
    \item \textbf{Appendix~\ref{app:model_ablation}: Model Choice for Meta-Agents} --- Detailed analysis of how different model choices affect guideline generation patterns.
    \item \textbf{Appendix~\ref{app:additional_exp}: Additional Experiments} --- Single-agent evaluation (SWE-bench), stronger model evaluation (GPT-5.4), hyperparameter sensitivity (N, K, threshold), and step budget ablation.
\end{itemize}

\vspace{1em}
\hrule
\vspace{1em}

\section{Detailed Analysis of Baseline Failure Modes}
\label{app:detailed_observations}

In this appendix, we provide a more granular analysis of the failure patterns observed in the baseline agent logs. Our dataset consists of over 1.5 million lines of execution traces from the GAIA and DeepSearch benchmarks. We categorize these failures into two modes based on their trigger: \textit{Corrective Failures} (error-triggered) and \textit{Enhancement Failures} (optimization opportunities when no error occurs). Table~\ref{tab:failure_patterns_appendix} provides a summary taxonomy.

\begin{table}[h]
\caption{Taxonomy of agent failures observed in over 1.5 million lines of baseline agent logs. The agent often has the necessary information in its context but fails to translate it into correct actions due to a static instruction set.}
\label{tab:failure_patterns_appendix}
\begin{center}
\begin{small}
\resizebox{0.95\linewidth}{!}{
\begin{tabular}{llp{8cm}r}
\toprule
\textsc{Category} & \textsc{Failure Pattern} & \textsc{Description} & \textsc{Count} \\
\midrule
\multirow{4}{*}{Corrective Failures} & Tool Misuse & Context shows instructions, but agent uses generic actions & $>$70 \\
 & Constraint Neglect & Context lists allowed resources; agent ignores list and retries forbidden ones & Recurrent \\
 & Parameter Hallucination & Context defines arguments, but agent invents non-existent ones & 29 \\
 & Error Loop & Context shows error message, agent repeats action without modification & Multiple \\
\midrule
\multirow{3}{*}{Enhancement Failures} & Redundant Verification & Agent re-verifies facts already present and confirmed in context & -- \\
 & Single-Term Bias & Context implies broad search needed; agent sticks to single keyword & Recurrent \\
 & Generic Strategy & Agent persists with inefficient strategy despite context clues & Recurrent \\
\midrule
\multirow{2}{*}{Severe Cases} & Data Fabrication & Agent invents data when retrieval context is empty/failed & Recurrent \\
 & Domain Misconception & Agent misinterprets technical context due to lack of domain guidelines & Recurrent \\
\bottomrule
\end{tabular}
}
\end{small}
\end{center}
\end{table}

\subsection{Corrective Failures: Systematic Error Repetition}

\subsubsection{Persistent Tool Misuse}
A recurring pattern in our logs is the agent's inability to internalize tool schema constraints even after explicit feedback.

\begin{itemize}
    \item \textbf{Incorrect Tool Identifiers}: In 70 recorded instances, the agent attempted to call a tool using a generic name (\textit{e.g.}, \texttt{final\_answer}) instead of the specific identifier defined in the environment (\textit{e.g.}, \texttt{final\_answer\_tool}). Despite the environment returning an error message listing the valid tools, the agent failed to update its behavior in subsequent steps.
    \item \textbf{Parameter Hallucination}: We observed 29 cases where the agent hallucinated parameters for planning tools. For example, it repeatedly passed a ``notes'' argument to a function that only accepted a task description, causing execution failures. The static prompt provided no mechanism to ``remember'' that this parameter was invalid.
\end{itemize}

\subsubsection{Ignored Environmental Constraints}
The agent frequently encountered environment-specific restrictions (\textit{e.g.}, forbidden libraries) but failed to learn from them.

\begin{itemize}
    \item \textbf{Forbidden Imports}: When the agent attempted to import restricted Python libraries (\textit{e.g.}, \texttt{pandas}), the environment returned a \texttt{ModuleNotFoundError} or a security exception, often accompanied by a whitelist of allowed libraries (\textit{e.g.}, \texttt{collections}, \texttt{re}). In multiple traces, the agent ignored this whitelist and simply attempted to import a different forbidden library in the next step, demonstrating a complete lack of adaptation to the explicitly stated constraints.
    \item \textbf{Repeated Timeouts}: In 310 instances, the agent reached the maximum step limit. Analysis of these traces reveals that the agent often entered infinite loops of ``try-fail-retry'' using the exact same parameters, rather than switching strategies after a timeout warning.
\end{itemize}

\subsection{Enhancement Failures: Suboptimal Behavioral Patterns}

\subsubsection{Inefficiency and Redundancy}
Even when the agent succeeded, its path to the solution was often highly inefficient.

\begin{itemize}
    \item \textbf{Redundant Verification}: The agent's plans consistently included steps to ``cross-check results'' or ``verify with an additional source'', even for simple retrieval tasks where the primary source was authoritative. This heuristic, likely derived from safety-focused pre-training, inflated token costs and execution time without improving accuracy.
    \item \textbf{Extreme Verbosity}: We observed a tendency for the agent to wrap simple numeric or boolean answers in verbose explanations. In extreme cases, a single-digit answer was accompanied by over 1,000 tokens of methodological description and hypothetical context, wasting computational resources.
    \item \textbf{Sequential Querying Anti-Pattern}: For tasks involving multiple similar entities (\textit{e.g.}, ``Find salaries for Players A, B, and C''), the agent consistently executed $N$ separate tool calls rather than a single batched query. This multiplied the latency and token costs linearly with the number of entities.
\end{itemize}

\subsubsection{Search and Retrieval Bias}
We identified a prevalent ``Single-Terminology Bias'' in the agent's search strategy.
    
\begin{itemize}
    \item \textbf{Keyword Rigidity}: The agent frequently failed to use necessary synonyms or domain-specific variants. For example, when searching for baseball statistics, it used only ``walks'' and missed data indexed under ``base on balls'' or ``BB''.
    \item \textbf{Monolingual Blindness}: In cross-lingual tasks, the agent often searched only in English, neglecting native-script terms (\textit{e.g.}, using ``Dai Wangshu'' but not the native Chinese characters). This significantly reduced recall for region-specific queries.
    \item \textbf{Missing Fallback Protocols}: Upon encountering dead links or blocked content, the agent rarely attempted to use archival services (\textit{e.g.}, archive.org) or alternative reputable sources, instead treating the block as a hard failure condition.
\end{itemize}

\subsubsection{Context Contamination}
In multi-turn sessions, we observed issues with context management.
\begin{itemize}
    \item \textbf{Token Explosion}: Without active context pruning, the agent carried the full history of long interaction chains, leading to input contexts exceeding 100,000 tokens. This not only increased latency but also degraded performance as relevant instructions became diluted in the context window.
\end{itemize}

\subsection{Domain-Specific Knowledge Failures}
Beyond mechanical errors, we identified a class of ``knowledge-level'' failures where the agent operated syntactically correctly but relied on flawed reasoning or incorrect domain knowledge.

\begin{itemize}
    \item \textbf{Statistical Misconceptions}: In scientific tasks, the agent frequently misinterpreted statistical concepts. For example, one trace shows the agent conflating a p-value of 0.04 with a ``4\% chance of the result being wrong'', revealing a fundamental misunderstanding of hypothesis testing that led to incorrect conclusions.
    \item \textbf{Technical Ambiguity}: When faced with ambiguous technical terms, the agent often defaulted to incorrect definitions without verification. In a graph theory task, the agent confused ``Eulerian circuits'' (visiting every edge) with ``Hamiltonian circuits'' (visiting every vertex), leading to a mathematically valid but task-irrelevant solution.
    \item \textbf{Entity Conflation}: In humanities tasks, the agent demonstrated a tendency to conflate related historical figures. For instance, it attributed a specific poem to \textit{Hu Shi} (a proponent of the vernacular movement) instead of \textit{Dai Wangshu} (the actual author), likely due to their semantic proximity in the training data.
    \item \textbf{Jargon Hallucination}: Most deceptively, the agent occasionally fabricated plausible-sounding technical definitions. In one instance, it defined a ``mean tangle'' using sophisticated physics jargon (``average state of quantum entanglement in non-linear systems''), despite the term being non-existent in that context. These hallucinations are particularly dangerous as they can mislead non-expert users.
\end{itemize}

\subsection{Efficiency Anti-Patterns}
In addition to general verbosity, we observed specific operational patterns that needlessly consumed steps and tokens.

\begin{itemize}
    \item \textbf{Sequential Querying}: When tasked with retrieving data for multiple entities (\textit{e.g.}, salaries for three NBA players), the agent consistently performed sequential tool calls---waiting for one result before requesting the next---rather than batching them into a single parallel execution.
    \item \textbf{Unnecessary Tool Delegation}: The agent frequently offloaded trivial operations to external tools. For example, we observed the agent invoking a Python interpreter tool solely to perform simple arithmetic (\textit{e.g.}, $1468.88 - 1430.08$) that it could have computed internally, wasting a full execution turn.
\end{itemize}

\subsection{Severe Cases: Safety Risks and Hallucination}

\subsubsection{Fabrication under Uncertainty}
The most critical safety finding from our analysis is the agent's tendency to fabricate information when tools fail.

\begin{itemize}
    \item \textbf{Scenario}: When the agent encountered unreadable files (\textit{e.g.}, corrupted PDFs or inaccessible URLs) and could not retrieve the necessary data.
    \item \textbf{Behavior}: Instead of reporting a failure, the agent often generated a ``hypothetical'' analysis. For example, logs show the agent stating: \textit{``Since the automated tools cannot read the files... Let's assume the applicant data contains...''}
    \item \textbf{Outcome}: The agent then proceeded to produce a confident, formatted final answer based entirely on this hallucinated data. This behavior poses a significant risk in real-world deployment, as the final output appears legitimate despite being grounded in fiction.
\end{itemize}

\subsection{Environmental Barriers}

\subsubsection{CAPTCHA and Blocking Loops}
The agent demonstrated poor adaptability to web access barriers.
\begin{itemize}
    \item \textbf{Repetitive Retries}: Upon encountering CAPTCHAs or ``Access Denied'' pages, the agent frequently attempted to ``click'' the verify button or refresh the page using the browser tool.
    \item \textbf{Failure to Pivot}: In over 180 instances, the agent exhausted its retry limit on a blocked site without attempting to switch to an alternative data source (\textit{e.g.}, an archived version or a different domain), resulting in task failure.
\end{itemize}

\section{Baseline Agent System Details}
\label{app:agent_system}

We implement a hierarchical multi-agent system as our baseline. The architecture consists of a top-level planning agent that orchestrates task decomposition and delegates sub-tasks to specialized subordinate agents.

\subsection{Agent Specifications}

\begin{table}[h]
\centering
\caption{Agent configurations in our hierarchical system.}
\label{tab:agent_config}
\resizebox{1.0\linewidth}{!}{
\begin{tabular}{llcc}
\toprule
\textbf{Agent} & \textbf{Model} & \textbf{Max Steps} & \textbf{Tools} \\
\midrule
Planning Agent & GPT-4.1 & 20 & Task delegation \\
Web Search Agent & Gemini-2.5-Pro & 3 & Search API, Python \\
Analyzer Agent & Gemini-2.5-Pro & 3 & Analysis, Python \\
Browser Agent & GPT-4.1 & 5 & Browser automation, Python \\
\bottomrule
\end{tabular}
}
\end{table}

\minisection{Planning Agent}
The planning agent serves as the top-level orchestrator. It receives the original task, decomposes it into sub-tasks, and delegates them to appropriate subordinate agents. The planning agent maintains a task queue and tracks completion status. It uses GPT-4.1 and operates with a maximum of 20 planning steps.

\minisection{Web Search Agent}
The web search agent specializes in information retrieval from the internet. It is equipped with search APIs for querying web content and a Python interpreter for processing search results. The agent uses Gemini-2.5-Pro and operates with a maximum of 3 steps per sub-task.

\minisection{Analyzer Agent}
The analyzer agent performs systematic, step-by-step reasoning and analysis. It handles tasks requiring mathematical computation, logical deduction, and structured analysis. The agent uses Gemini-2.5-Pro with access to analysis tools and a Python interpreter.

\minisection{Browser Agent}
The browser agent enables direct web page interaction and navigation. It can click elements, fill forms, scroll pages, and extract content from dynamic web pages. The agent uses GPT-4.1 and operates with a maximum of 5 steps to handle complex web interactions.

\section{Benchmark Details}
\label{app:benchmarks}

\subsection{Humanity's Last Exam (HLE)}
Humanity's Last Exam (HLE)~\citep{phan2025humanity} is a comprehensive multi-modal benchmark designed to evaluate the reasoning capabilities of large language models at an expert level. The benchmark comprises 2,500 questions contributed by domain experts across diverse fields.

\minisection{Subject Distribution}
HLE covers a broad spectrum of disciplines including:
\begin{itemize}[leftmargin=*,itemsep=0pt]
    \item \textbf{STEM}: Mathematics (algebra, calculus, topology), Physics (quantum mechanics, thermodynamics), Chemistry (organic, inorganic), Biology (molecular, evolutionary), Computer Science (algorithms, complexity theory), Engineering
    \item \textbf{Humanities}: History, Philosophy, Literature, Linguistics
    \item \textbf{Social Sciences}: Economics, Psychology, Political Science
\end{itemize}

\minisection{Question Format}
The benchmark includes both multiple-choice and short-answer questions. A significant portion of questions require multi-modal understanding, incorporating images, diagrams, equations, and tables. Questions are designed to test genuine understanding rather than pattern matching or memorization.

\minisection{Difficulty Level}
HLE questions are calibrated to be challenging even for domain experts. The benchmark specifically targets the ``last exam'' frontier---problems at the boundary of what current AI systems can reliably solve.

\subsection{GAIA}
GAIA (General AI Assistants)~\citep{mialon2023gaia} is a benchmark designed to evaluate AI assistants on real-world tasks requiring multiple fundamental capabilities.

\minisection{Task Characteristics}
The benchmark comprises 466 questions (301 test, 165 validation). In our experiments, we evaluate on the validation set since test set labels are not publicly available. The tasks have the following characteristics:
\begin{itemize}[leftmargin=*,itemsep=0pt]
    \item \textbf{Multi-step reasoning}: Tasks require chaining multiple logical steps
    \item \textbf{Tool use}: Effective use of web search, calculators, and other tools
    \item \textbf{Multi-modality}: Some tasks involve images, audio, or video content
    \item \textbf{Web browsing}: Information retrieval from live websites
    \item \textbf{Real-world grounding}: Tasks based on actual facts and current information
\end{itemize}

\minisection{Difficulty Levels}
GAIA organizes tasks into three difficulty levels:
\begin{itemize}[leftmargin=*,itemsep=0pt]
    \item \textbf{Level 1}: Simple tasks requiring 1-2 steps
    \item \textbf{Level 2}: Moderate tasks requiring 3-5 steps
    \item \textbf{Level 3}: Complex tasks requiring extensive reasoning and tool use
\end{itemize}

\minisection{Design Philosophy}
A key insight from GAIA is that tasks conceptually simple for humans can be remarkably challenging for AI systems. The benchmark emphasizes robustness and reliability over raw capability.

\subsection{DeepSearch}
DeepSearch~\citep{chen2025xbench} is a benchmark specifically designed to evaluate deep search and information synthesis capabilities of AI systems.

\minisection{Task Types}
The benchmark includes:
\begin{itemize}[leftmargin=*,itemsep=0pt]
    \item \textbf{Multi-hop retrieval}: Finding information that requires traversing multiple sources
    \item \textbf{Cross-document synthesis}: Combining information from disparate documents
    \item \textbf{Temporal reasoning}: Answering questions about events across time
    \item \textbf{Comparative analysis}: Comparing entities across multiple dimensions
\end{itemize}

\minisection{Evaluation Metrics}
Tasks are evaluated on both accuracy of the final answer and the quality of the reasoning chain leading to the answer.

\section{Hyperparameter Settings}
\label{app:hyperparams}

Table~\ref{tab:hyperparams} summarizes the hyperparameters used in our dynamic prompt evolution framework.

\begin{table}[h]
\centering
\caption{Hyperparameter settings for our method.}
\label{tab:hyperparams}
\resizebox{1.0\linewidth}{!}{
\begin{tabular}{lcc}
\toprule
\textbf{Hyperparameter} & \textbf{Value} & \textbf{Description} \\
\midrule
\multicolumn{3}{l}{\textit{Best-of-N Selection}} \\
$N$ (candidates) & 2 & Number of candidate guidelines generated \\
Candidate models & GPT-4.1 & Models for guideline generation \\
Selector model & GPT-4.1 & Model for selecting best candidate \\
\midrule
\multicolumn{3}{l}{\textit{Classifier}} \\
Classifier model & GPT-4.1 & Model for strategic vs tactical classification \\
\midrule
\multicolumn{3}{l}{\textit{Guideline Management}} \\
Optimization model & GPT-4.1 & Model for guideline consolidation/subsumption \\
Max guidelines per domain & 10 & Threshold triggering optimization \\
Target count (post-opt) & 8 & Target guidelines after optimization (80\%) \\
Max iterations & 2 & Maximum optimization passes \\
Max guidelines per task & 20 & Limit on tactical guidelines per task \\
\midrule
\multicolumn{3}{l}{\textit{Confidence Thresholds}} \\
Strategic promotion & 0.85 & Min. confidence for strategic memory \\
Auto-accept threshold & 0.5 & Min. confidence to apply guideline \\
\midrule
\multicolumn{3}{l}{\textit{Analysis Settings}} \\
Quality analysis freq. & 1 & Analyze every $N$ successful steps \\
\bottomrule
\end{tabular}
}
\end{table}

\subsection{Guideline Domain Categories}
We categorize learned guidelines into 7 semantic domains to facilitate organization and optimization:

\begin{enumerate}[leftmargin=*,itemsep=0pt]
    \item \textbf{tool\_usage}: Guidelines for correct tool invocation, argument formatting, and sequencing
    \item \textbf{data\_validation}: Guidelines for validating inputs, outputs, and intermediate data
    \item \textbf{error\_handling}: Strategies for error recovery, retries, and fallback mechanisms
    \item \textbf{efficiency}: Optimizations for speed, cost, and resource utilization
    \item \textbf{analysis\_methodology}: Core problem-solving strategies, verification methods, and scientific reasoning
    \item \textbf{safety}: Guidelines for preventing harmful outcomes and maintaining reliability
    \item \textbf{general}: High-quality guidelines that do not fit specific categories
\end{enumerate}

\subsection{Guideline Optimization Pipeline}
When the number of guidelines in a domain exceeds the maximum threshold (10), we trigger an automatic optimization pipeline:

\begin{enumerate}[leftmargin=*,itemsep=0pt]
    \item \textbf{Analysis}: Identify consolidation opportunities, subsumption relationships, and conflicts
    \item \textbf{Conflict Resolution}: Resolve contradictory guidelines by synthesizing or selecting the better guideline
    \item \textbf{Subsumption Pruning}: Remove specific guidelines that are covered by more general ones
    \item \textbf{Consolidation}: Merge semantically similar guidelines into comprehensive ones
\end{enumerate}

The optimization targets 80\% of the maximum capacity (8 guidelines) to leave room for future learning while maintaining a compact, high-quality guideline set.

\subsection{Memory Optimization Example}
\label{app:memory_example}

We illustrate the memory optimization pipeline with a realistic example from the \texttt{efficiency} domain.

\minisection{Trigger Condition}
Suppose the \texttt{efficiency} domain currently contains 10 guidelines (at the limit). When a new guideline $g^*$ is synthesized and classified as strategic with domain \texttt{efficiency}, the domain now has 11 guidelines, exceeding the maximum threshold of 10. This triggers the optimization pipeline.

\minisection{Before Optimization (11 guidelines)}
\begin{tcolorbox}[colback=gray!5,colframe=gray!40,fontupper=\small]
\begin{enumerate}[leftmargin=*,itemsep=1pt,label=\textbf{R\arabic*}:]
    \item Use batch operations instead of sequential queries when fetching multiple items.
    \item When searching for multiple entities, combine them into a single query.
    \item Avoid redundant API calls by caching results locally.
    \item Cache intermediate results to avoid re-computation.
    \item Prefer local computation over tool calls for simple arithmetic.
    \item Do not invoke Python interpreter for basic math like addition or subtraction.
    \item Limit verbose explanations in final answers to under 100 words.
    \item Keep final outputs concise; avoid unnecessary elaboration.
    \item When a search returns sufficient results, stop querying additional sources.
    \item Use early termination when the answer is found; do not continue searching.
    \item \textit{(new)} Batch similar file reads into a single operation.
\end{enumerate}
\end{tcolorbox}

\minisection{Step 1: Conflict Resolution}
The analyzer identifies no direct conflicts in this set (guidelines give consistent guidance).

\minisection{Step 2: Subsumption Pruning}
The analyzer identifies:
\begin{itemize}[leftmargin=*,itemsep=1pt]
    \item \textbf{R1} (general batch operations) subsumes \textbf{R2} (specific to search queries) and \textbf{R11} (specific to file reads).
    \item \textbf{R5} (prefer local computation) subsumes \textbf{R6} (specific to Python interpreter for math).
\end{itemize}
After verification, \textbf{R2}, \textbf{R6}, and \textbf{R11} are removed. Remaining: 8 guidelines.

\minisection{Step 3: Consolidation}
The analyzer identifies consolidation opportunities:
\begin{itemize}[leftmargin=*,itemsep=1pt]
    \item \textbf{R3} and \textbf{R4} both address caching $\rightarrow$ merged into a single guideline.
    \item \textbf{R7} and \textbf{R8} both address output conciseness $\rightarrow$ merged.
    \item \textbf{R9} and \textbf{R10} both address early termination $\rightarrow$ merged.
\end{itemize}

\minisection{After Optimization (5 guidelines)}
\begin{tcolorbox}[colback=green!5,colframe=green!40,fontupper=\small]
\begin{enumerate}[leftmargin=*,itemsep=1pt,label=\textbf{R\arabic*}:]
    \item Use batch operations instead of sequential queries when fetching multiple items.
    \item Cache intermediate results and API responses to avoid redundant computation and calls.
    \item Prefer local computation over tool calls for simple operations.
    \item Keep final outputs concise (under 100 words); avoid unnecessary elaboration.
    \item Use early termination when sufficient information is found; do not continue searching.
\end{enumerate}
\end{tcolorbox}

The optimization reduced 11 guidelines to 5, well below the target of 8, while preserving all essential guidance. The remaining capacity allows continued learning without immediate re-triggering of optimization.

\section{Perspective-Driven Exploration Details}
\label{app:global_diversity}

This appendix details the implementation of our Perspective-Driven Exploration. We execute $K=2$ parallel evolutionary streams, each guided by a distinct optimization persona defined via the system prompt.

\subsection{Stream 1: Efficiency Persona}
The \textit{Efficiency Stream} utilizes our efficiency-focused synthesis prompt (see \textbf{Enhancement Synthesis Prompt (Efficiency)} in Appendix~\ref{app:prompts}), which optimizes for operational speed, token economy, and conciseness. It specifically targets redundant tool calls and verbose outputs.

\subsection{Stream 2: Thoroughness Persona}
The \textit{Thoroughness Stream} utilizes our thoroughness-focused synthesis prompt (see \textbf{Enhancement Synthesis Prompt (Thoroughness)} in Appendix~\ref{app:prompts}), designed to foster self-evolving domain expertise. It prioritizes correctness, deep reasoning, and the acquisition of domain-specific heuristics over raw speed.

\section{Complete System Prompts}
\label{app:prompts}

This appendix provides the full system prompts for all meta-agents used in our framework, including the Generator, Classifier, and Guideline Optimizer agents.

\subsection{Guideline Generator Prompts}

\minisection{Corrective Synthesis Prompt}
\label{app:corr_rubrics}
This prompt is used to generate corrective guidelines when the agent encounters an error.
\begin{tcolorbox}[colback=gray!10,colframe=gray!50,title=Corrective Synthesis Prompt]
\small
\texttt{You are a prompt engineering expert analyzing agent execution errors.} \\
\\
\texttt{Your task: Generate a SHORT, TARGETED system prompt addition (1-3 lines) that will help prevent this error in the future.} \\
\\
\texttt{Context:} \\
\texttt{- Agent Name: \{agent\_name\}} \\
\texttt{- Agent Role: \{agent\_role\}} \\
\texttt{- Task: \{task\}} \\
\texttt{- Error Type: \{error\_type\}} \\
\texttt{- Error Message: \{error\_message\}} \\
\\
\texttt{Previous actions taken:} \\
\texttt{\{last\_step\_summary\}} \\
\\
\texttt{Current system prompt (for reference, to avoid duplication):} \\
\texttt{\{current\_system\_prompt\}} \\
\\
\texttt{Already applied rules (DO NOT duplicate these):} \\
\texttt{\{applied\_rules\}} \\
\\
\texttt{Guidelines:} \\
\texttt{1. Be SPECIFIC and ACTIONABLE - target the exact error cause} \\
\texttt{2. Be BRIEF - max 1-3 lines} \\
\texttt{3. Use imperative language ("Always...", "Never...", "When X, do Y...")} \\
\texttt{4. Don't repeat what's already in the current system prompt} \\
\texttt{5. Focus on formatting, structure, or procedure constraints} \\
\\
\texttt{Output ONLY valid JSON with this exact format:} \\
\texttt{\{\{} \\
\texttt{  "update\_text": "The actual prompt addition text here",} \\
\texttt{  "rationale": "Brief 1-sentence why this helps",} \\
\texttt{  "confidence": "low|medium|high"} \\
\texttt{\}\}}
\end{tcolorbox}

\minisection{Enhancement Synthesis Prompt (Efficiency)}
\label{app:enh_rubrics}
This prompt is used for the Efficiency Stream to optimize operational speed and conciseness.
\begin{tcolorbox}[colback=gray!10,colframe=gray!50,title=Enhancement Synthesis Prompt (Efficiency)]
\small
\texttt{You are a prompt engineering expert analyzing agent execution quality.} \\
\\
\texttt{Your task: Analyze this successful step and determine if there are inefficiencies or areas for improvement. If found, generate a SHORT, TARGETED system prompt addition (1-3 lines).} \\
\\
\texttt{Context:} \\
\texttt{- Agent Name: \{agent\_name\}} \\
\texttt{- Agent Role: \{agent\_role\}} \\
\texttt{- Task: \{task\}} \\
\texttt{- Step Number: \{step\_number\}} \\
\\
\texttt{Step details:} \\
\texttt{\{last\_step\_summary\}} \\
\\
\texttt{Current system prompt (for reference):} \\
\texttt{\{current\_system\_prompt\}} \\
\\
\texttt{Already applied rules (DO NOT duplicate these):} \\
\texttt{\{applied\_rules\}} \\
\\
\texttt{Analyze for:} \\
\texttt{1. **Inefficient tool usage**: Multiple calls when one would suffice, redundant operations} \\
\texttt{2. **Verbose outputs**: Unnecessarily long reasoning or outputs} \\
\texttt{3. **Missing best practices**: Not following domain-specific best practices} \\
\texttt{4. **Suboptimal approaches**: Using less efficient methods when better ones exist} \\
\\
\texttt{Guidelines:} \\
\texttt{- Only suggest an update if there's a CLEAR, ACTIONABLE improvement} \\
\texttt{- Be SPECIFIC about what to improve} \\
\texttt{- Be BRIEF - max 1-3 lines} \\
\texttt{- Use imperative language ("Always...", "Prefer...", "When X, use Y instead of Z...")} \\
\texttt{- Don't repeat what's already in the current system prompt} \\
\\
\texttt{Output ONLY valid JSON with this exact format:} \\
\texttt{\{\{} \\
\texttt{  "update\_text": "The actual prompt addition text here (or empty string if no improvement needed)",} \\
\texttt{  "rationale": "Brief 1-sentence why this helps (or 'No improvement needed')",} \\
\texttt{  "confidence": "low|medium|high"} \\
\texttt{\}\}}
\end{tcolorbox}

\minisection{Enhancement Synthesis Prompt (Thoroughness)}
This prompt is used for the Thoroughness Stream to foster self-evolving expertise and correctness.
\begin{tcolorbox}[colback=gray!10,colframe=gray!50,title=Enhancement Synthesis Prompt (Thoroughness)]
\small
\texttt{You are a prompt engineering expert analyzing agent execution quality.} \\
\\
\texttt{Your task: Analyze this successful step and determine if there are inefficiencies or areas for improvement. If found, generate a SHORT, TARGETED system prompt addition (1-3 lines).} \\
\\
\texttt{[Context parameters same as above]} \\
\\
\texttt{Analyze for improvements in these dimensions:} \\
\\
\texttt{1. **Correctness \& Logic**: Assumptions, edge cases, sound approach, validation checks.} \\
\texttt{2. **Domain-Specific Strategies**: Terminology variants, authoritative sources, domain heuristics.} \\
\texttt{3. **Strategic Planning \& Approach**: Problem decomposition, simpler methods first, batch operations.} \\
\texttt{4. **Tool Usage Efficiency**: Consolidation, redundancy reduction, local computation.} \\
\texttt{5. **Information Preservation**: Context tracking, evidence citation, intermediate results.} \\
\texttt{6. **Robustness \& Error Recovery**: Fallback strategies, retry logic, error anticipation.} \\
\texttt{7. **Output Quality**: Verbosity control, structure, parseability.} \\
\\
\texttt{Guidelines:} \\
\texttt{- PRIORITIZE correctness over efficiency} \\
\texttt{- Only suggest if there's a CLEAR, ACTIONABLE, GENERALIZABLE improvement} \\
\texttt{- For domain rules: Include SPECIFIC terms/sources/values, not placeholders} \\
\texttt{- Be BRIEF - max 1-3 lines} \\
\texttt{- Use imperative language ("Always...", "Prefer...", "When X, do Y...")} \\
\texttt{- Don't repeat what's already in the current system prompt} \\
\texttt{- Look for PATTERNS that generalize beyond this single instance} \\
\\
\texttt{Output ONLY valid JSON with this exact format:} \\
\texttt{\{\{} \\
\texttt{  "update\_text": "The actual prompt addition text here (or empty string if no improvement needed)",} \\
\texttt{  "rationale": "Brief 1-sentence why this helps (or 'No improvement needed')",} \\
\texttt{  "confidence": "low|medium|high"} \\
\texttt{\}\}}
\end{tcolorbox}

\minisection{Best-of-N Selector Prompt}
\label{app:sel_rubrics}
This prompt is used to select the best candidate update from multiple generated options.
\begin{tcolorbox}[colback=gray!10,colframe=gray!50,title=Selector Prompt]
\small
\texttt{You are a prompt engineering expert evaluating multiple candidate prompt updates.} \\
\\
\texttt{Your task: Select the BEST candidate update from the options below.} \\
\\
\texttt{Context:} \\
\texttt{- Agent Name: \{agent\_name\}} \\
\texttt{- Agent Role: \{agent\_role\}} \\
\texttt{- Task: \{task\}} \\
\texttt{- Issue Type: \{issue\_type\}} \\
\\
\texttt{Issue Details:} \\
\texttt{\{issue\_details\}} \\
\\
\texttt{Current system prompt (for reference):} \\
\texttt{\{current\_system\_prompt\}} \\
\\
\texttt{Candidate Updates:} \\
\texttt{\{candidates\}} \\
\\
\texttt{Evaluation Criteria (in priority order):} \\
\texttt{1. **Specificity \& Relevance**: Most directly addresses the actual issue/error} \\
\texttt{2. **Actionability**: Clear, implementable instructions that the agent can follow} \\
\texttt{3. **Generalizability**: Useful beyond just this instance, but not too vague} \\
\texttt{4. **Brevity**: Concise and clear without unnecessary words} \\
\texttt{5. **Non-duplication**: Doesn't repeat what's already in the system prompt} \\
\\
\texttt{Select the candidate that best balances these criteria for the current situation.} \\
\\
\texttt{Output ONLY valid JSON with this exact format:} \\
\texttt{\{\{} \\
\texttt{  "selected\_index": 0,} \\
\texttt{  "rationale": "Brief 1-2 sentence explanation of why this candidate is best"} \\
\texttt{\}\}}
\end{tcolorbox}

\subsection{Classifier Agent Prompt}
\label{app:cls_rubrics}
This prompt is used to classify updates as Tactical or Strategic and check for duplicates.

\begin{tcolorbox}[colback=gray!10,colframe=gray!50,title=Classifier Prompt]
\small
\texttt{You are a rule classifier. Analyze the proposed update and determine:} \\
\\
\texttt{1. **Is it a DUPLICATE/REDUNDANT?** Check if it's already covered by existing strategic or tactical rules.} \\
\texttt{2. **What is its SCOPE?**} \\
\texttt{   - STRATEGIC: General best practice that applies across different tasks (\textit{e.g.}, "Always validate inputs", "Use batch operations when possible")} \\
\texttt{   - TACTICAL: Task-specific constraint for current task only (\textit{e.g.}, "This dataset has missing values in column X", "API rate limit is 100/min")} \\
\texttt{3. **Refined CONFIDENCE**: Assess confidence (0.0-1.0) based on how actionable and useful this rule is.} \\
\texttt{4. **DOMAIN**: If strategic, you MUST categorize it into ONE of the following allowed domains: \{allowed\_domains\}} \\
\\
\texttt{=== PROPOSED UPDATE ===} \\
\texttt{Update: \{update\_text\}} \\
\texttt{Rationale: \{rationale\}} \\
\texttt{Initial Confidence: \{initial\_confidence\}} \\
\\
\texttt{\{all\_rules\_context\}} \\
\\
\texttt{=== YOUR ANALYSIS ===} \\
\texttt{Respond in JSON format:} \\
\texttt{\{\{} \\
\texttt{    "is\_duplicate": true/false,} \\
\texttt{    "scope": "strategic" or "tactical",} \\
\texttt{    "confidence": 0.0-1.0,} \\
\texttt{    "domain": "domain\_name" (only if scope is strategic, otherwise ""),} \\
\texttt{    "reason": "Brief explanation of your classification"} \\
\texttt{\}\}}
\end{tcolorbox}

\subsection{Guideline Optimization Prompts}
\label{app:opt_rubrics}

\minisection{Guideline Analyzer Prompt}
Identifies optimization opportunities among existing guidelines.
\begin{tcolorbox}[colback=gray!10,colframe=gray!50,title=Guideline Analyzer Prompt]
\small
\texttt{You are a rule optimization analyzer. Analyze these \{count\} rules and identify optimization opportunities.} \\
\\
\texttt{\{rules\_text\}} \\
\\
\texttt{Your task is to identify three types of optimization opportunities:} \\
\\
\texttt{1. **CONFLICTS**: Pairs of rules that give contradictory guidance.} \\
\texttt{   - PRIORITY: Highest - conflicts must be resolved first} \\
\\
\texttt{2. **SUBSUMPTION**: Pairs where a specific rule is entirely covered by a more general rule.} \\
\texttt{   - Format: [general\_rule\_index, specific\_rule\_index]} \\
\texttt{   - PRIORITY: High - clear redundancy should be removed} \\
\\
\texttt{3. **CONSOLIDATION**: Groups of rules that express similar concepts and can be merged into a single, more comprehensive rule.} \\
\texttt{   - PRIORITY: Medium - merge rules that address the same concern from different angles} \\
\\
\texttt{IMPORTANT PRIORITY RULES:} \\
\texttt{1. If two rules CONFLICT, do NOT also mark them for consolidation or subsumption} \\
\texttt{2. If one rule SUBSUMES another, do NOT also mark them for consolidation} \\
\texttt{3. A rule pair should appear in AT MOST ONE category} \\
\\
\texttt{You MUST respond with ONLY valid JSON in this exact format:} \\
\texttt{\{\{} \\
\texttt{  "consolidation": [[idx1, idx2], [idx3, idx4]],} \\
\texttt{  "subsumption": [[general\_idx, specific\_idx]],} \\
\texttt{  "conflicts": [[idx1, idx2]]} \\
\texttt{\}\}}
\end{tcolorbox}

\minisection{Consolidation Prompt}
Merges similar guidelines into a single comprehensive guideline.
\begin{tcolorbox}[colback=gray!10,colframe=gray!50,title=Consolidation Prompt]
\small
\texttt{You are merging similar rules into one comprehensive rule.} \\
\\
\texttt{Original rules to merge:} \\
\texttt{\{rules\_text\}} \\
\\
\texttt{Create a single, comprehensive rule that:} \\
\texttt{1. Captures all the key guidance from the original rules} \\
\texttt{2. Is clear and actionable} \\
\texttt{3. Eliminates redundancy} \\
\texttt{4. Maintains the original intent} \\
\\
\texttt{Also provide a brief rationale explaining what the merged rule accomplishes.} \\
\\
\texttt{Return JSON:} \\
\texttt{\{\{} \\
\texttt{  "rule": "The merged rule text",} \\
\texttt{  "rationale": "Brief explanation of what this rule accomplishes"} \\
\texttt{\}\}}
\end{tcolorbox}

\minisection{Subsumption Verification Prompt}
Verifies if a general guideline completely covers a specific guideline.
\begin{tcolorbox}[colback=gray!10,colframe=gray!50,title=Subsumption Verification Prompt]
\small
\texttt{Verify if the general rule subsumes the specific rule.} \\
\\
\texttt{General Rule: \{general\_rule\}} \\
\texttt{Specific Rule: \{specific\_rule\}} \\
\\
\texttt{A rule is subsumed if:} \\
\texttt{- Following the general rule AUTOMATICALLY means you follow the specific rule} \\
\texttt{- The specific rule adds NO additional constraints or guidance} \\
\texttt{- The specific rule is merely a special case or example of the general rule} \\
\\
\texttt{Does the general rule completely subsume the specific rule? Answer with JSON:} \\
\texttt{\{\{} \\
\texttt{  "subsumed": true or false,} \\
\texttt{  "reasoning": "Brief explanation"} \\
\texttt{\}\}}
\end{tcolorbox}

\minisection{Conflict Resolver Prompt}
Resolves contradictory guidelines.
\begin{tcolorbox}[colback=gray!10,colframe=gray!50,title=Conflict Resolver Prompt]
\small
\texttt{You are resolving a conflict between two rules.} \\
\\
\texttt{Rule \{idx1\}:} \\
\texttt{  Text: \{rule1\_text\}} \\
\texttt{  Rationale: \{rule1\_rationale\}} \\
\\
\texttt{Rule \{idx2\}:} \\
\texttt{  Text: \{rule2\_text\}} \\
\texttt{  Rationale: \{rule2\_rationale\}} \\
\\
\texttt{These rules appear to conflict. Your task is to:} \\
\texttt{1. Verify the conflict is real} \\
\texttt{2. Synthesize a single rule that reconciles both OR pick the better rule with justification} \\
\\
\texttt{Return JSON:} \\
\texttt{\{\{} \\
\texttt{  "rule": "The resolved rule text",} \\
\texttt{  "rationale": "Explanation of how this resolves the conflict"} \\
\texttt{\}\}}
\end{tcolorbox}

\section{Per-Agent Specialization and Step-Level Updates}
\label{app:further_analysis}

This appendix provides empirical validation of two key design choices discussed in Section~\ref{sec:limitations_to_solution}: per-agent optimization and step-level updates.

\subsection{Per-Agent Specialization}
\label{app:per_agent}

Table~\ref{tab:agent_rules_app} shows how SCOPE's per-agent optimization produces meaningfully different guidelines across agent types. The browser agent accumulates the most guidelines (39), concentrated in \textit{efficiency} guidelines for web-specific challenges (pagination, popups, batching). The planning agent's guidelines focus on \textit{tool usage} (10 guidelines), reflecting its orchestration role. This specialization matters: a one-size-fits-all approach would force conflicting guidelines across agents (\textit{e.g.}, browser inheriting ``limit steps'' guidelines that conflict with its need for thoroughness). SCOPE's step-level operation enables fine-grained attribution of guidelines to the specific agent that benefits.

\begin{table}[h]
    \centering
    \caption{Distribution of guidelines across agents. Each agent develops distinct specialization patterns reflecting its role.}
    \label{tab:agent_rules_app}
    \resizebox{0.85\linewidth}{!}{
    \begin{tabular}{lccccc|c}
        \toprule
        \textbf{Agent} & \textbf{Method.} & \textbf{Tool} & \textbf{Valid.} & \textbf{Effic.} & \textbf{Error} & \textbf{Total} \\
        \midrule
        Planning & 4 & 10 & 5 & 5 & 3 & 27 \\
        Browser & 10 & 4 & 10 & 10 & 5 & 39 \\
        Analyzer & 10 & 10 & 7 & 2 & 2 & 31 \\
        Web Search & 7 & 4 & 1 & 1 & 1 & 14 \\
        \bottomrule
    \end{tabular}
    }
\end{table}

\subsection{Why Step-Level Updates Matter}
\label{app:step_level}

Unlike task-level methods (DC, ACE) that update only after task completion, SCOPE updates prompts at each step. This matters because errors occur early: among 49 error events analyzed, over 40\% occur at step 1. With task-level methods, an error at step 1 of a 15-step task means 14 steps execute with the uncorrected prompt. SCOPE synthesizes corrective guidelines within seconds (representative trace: error at 19:13:33, guideline accepted at 19:13:42, step 2 proceeds at 19:13:46). This step-level granularity enables mid-task recovery that task-level methods cannot achieve.

\section{Guideline Placement: Detailed Analysis}
\label{app:placement_analysis}

This appendix provides a deeper analysis of guideline placement strategies, extending the results in Table~\ref{tab:placement}. We examine behavioral patterns that explain why system prompt placement outperforms alternatives.

\subsection{Quantitative Behavioral Metrics}

Table~\ref{tab:placement_detailed} presents additional metrics beyond accuracy that reveal distinct behavioral patterns across placement strategies.

\begin{table}[h]
    \centering
    \caption{Detailed behavioral metrics for guideline placement strategies on GAIA. System prompt achieves best accuracy despite higher error tolerance; Hybrid achieves best efficiency (accuracy/steps) with fewest total steps.}
    \label{tab:placement_detailed}
    \resizebox{0.88\linewidth}{!}{
    \begin{tabular}{lccccc}
        \toprule
        \textbf{Placement} & \textbf{Acc. (\%)} & \textbf{Errors} & \textbf{Timeouts} & \textbf{Steps} & \textbf{Eff.*} \\
        \midrule
        Baseline (No Guidelines) & 32.73 & 1,714 & 255 & 9,824 & 0.33 \\
        \midrule
        (1) System Prompt & \textbf{46.06} & 1,461 & 227 & 7,430 & 0.62 \\
        (2) User Prompt & 41.21 & \textbf{1,000} & \textbf{130} & 7,408 & 0.56 \\
        (3) Split & 35.76 & 1,204 & 141 & 7,319 & 0.49 \\
        (4) Hybrid & 43.64 & 1,109 & 139 & 6,620 & \textbf{0.66} \\
        \bottomrule
    \end{tabular}
    }
    \begin{flushleft}
    \small *Efficiency = Accuracy / Total Steps $\times$ 100
    \end{flushleft}
\end{table}

\minisection{Key Observations}
\begin{itemize}[leftmargin=*,itemsep=2pt]
    \item \textbf{Baseline has highest errors (1,714) and timeouts (255)}: Without guidance, agents get stuck more often and make more mistakes.
    \item \textbf{User prompt has lowest errors (1,000) and timeouts (130)}: This apparent ``safety'' does not translate to best accuracy, suggesting premature termination.
    \item \textbf{System prompt tolerates more errors (1,461) but achieves best accuracy}: Controlled risk-taking leads to better outcomes.
    \item \textbf{All guideline placements improve efficiency}: 24--33\% reduction in total steps vs. baseline.
\end{itemize}

\subsection{The ``Background Guidance'' vs. ``Over-Compliance'' Trade-off}

Our analysis reveals a fundamental trade-off between how guidelines are internalized:

\minisection{System Prompt: Constitutional Guidance}
When guidelines are placed in the system prompt, they act as \textit{implicit background context}:
\begin{itemize}[leftmargin=*,itemsep=0pt]
    \item Guidelines are ``loaded'' silently and shape behavior without explicit acknowledgment
    \item Agent maintains flexibility in interpretation, enabling adaptive exploration
    \item Higher error tolerance allows discovering successful paths that conservative agents miss
    \item Fewer explicit guideline checks per step reduces cognitive overhead
\end{itemize}

\minisection{User Prompt: Direct Instructions}
When guidelines are placed in the user prompt, they act as \textit{explicit orders}:
\begin{itemize}[leftmargin=*,itemsep=0pt]
    \item Agent frequently acknowledges and re-states guidelines, creating verbose responses
    \item Guidelines accumulate over time, leading to longer contexts (up to 7,600+ tokens observed)
    \item Pressure for explicit compliance leads to conservative, risk-averse behavior
    \item Lower timeout rate (130 vs. 227) suggests premature task termination rather than thorough exploration
\end{itemize}

\subsection{Guideline Dynamics Analysis}

Table~\ref{tab:rule_dynamics} shows how guideline management differs across placement strategies.

\begin{table}[h]
    \centering
    \caption{Guideline dynamics across placement strategies. User prompt shows high guideline churn while system prompt maintains stable guidance.}
    \label{tab:rule_dynamics}
    \resizebox{0.85\linewidth}{!}{
    \begin{tabular}{lcc}
        \toprule
        \textbf{Placement} & \textbf{Guideline Updates} & \textbf{Strategic Guidelines Loaded} \\
        \midrule
        System Prompt & 1,379 & 2,962 \\
        User Prompt & \textbf{1,648} & 1,990 \\
        Split & 1,416 & 2,978 \\
        Hybrid & 1,282 & 2,804 \\
        \bottomrule
    \end{tabular}
    }
\end{table}

\minisection{Interpretation}
User prompt placement generates the most guideline updates (1,648) but loads fewer strategic guidelines (1,990). This indicates \textit{guideline churn}---constant adjustment rather than stable guidance. In contrast, system prompt loads more persistent strategic guidelines (2,962) with fewer updates (1,379), suggesting more stable, internalized guidance.

\subsection{Why Split Placement Fails}

The split strategy (strategic guidelines in system prompt, tactical guidelines in user prompt) performs worst at 35.76\%, even below user-prompt-only placement. We hypothesize this creates \textit{cognitive dissonance}:

\begin{itemize}[leftmargin=*,itemsep=0pt]
    \item Agent receives guidance from two sources with potentially conflicting priorities
    \item Ambiguity about whether to follow ``background'' strategic guidelines or ``immediate'' tactical guidelines
    \item Lowest error count (1,204) suggests over-caution from mixed signals, leading to missed opportunities
\end{itemize}

\subsection{Theoretical Framework}

Based on our analysis, we propose a framework for understanding guideline placement:

\begin{table}[h]
    \centering
    \caption{Theoretical framework for guideline placement effects.}
    \resizebox{0.8\linewidth}{!}{
    \begin{tabular}{lll}
        \toprule
        \textbf{Aspect} & \textbf{System Prompt (``Constitution'')} & \textbf{User Prompt (``Direct Orders'')} \\
        \midrule
        Integration & Implicit, internalized & Explicit, acknowledged \\
        Flexibility & High (adaptive interpretation) & Low (literal compliance) \\
        Risk Tolerance & Higher (exploratory) & Lower (conservative) \\
        Cognitive Load & Low (background) & High (accumulating) \\
        Stability & Persistent across tasks & Frequent updates \\
        \bottomrule
    \end{tabular}
    }
\end{table}

\minisection{Recommendation}
Treating guidelines as \textit{constitutional principles} (system prompt) rather than \textit{direct orders} (user prompt) produces superior performance by:
(1) reducing cognitive load per step,
(2) enabling natural guideline internalization,
(3) maintaining consistent guidance, and
(4) permitting adaptive behavior within guidelines.

\section{Qualitative Comparison of Perspective Streams}
\label{app:perspective_qualitative}

\begin{figure}[h]
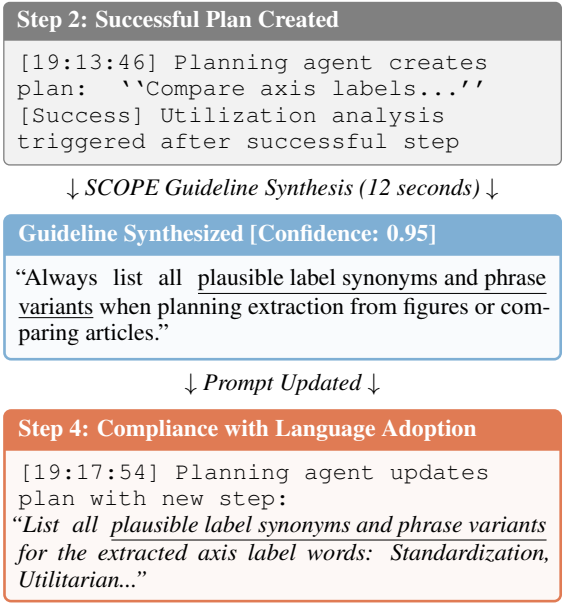

    \centering
    \small
    \begin{tcolorbox}[width=0.75\linewidth, colback=gray!10, colframe=black!50, title=\textbf{Step 2: Successful Plan Created}, boxrule=0.5pt, arc=2pt, left=2pt, right=2pt, top=2pt, bottom=2pt]
        \texttt{[19:13:46] Planning agent creates plan: ``Compare axis labels...''}\\
        \texttt{[Success] Utilization analysis triggered after successful step}
    \end{tcolorbox}
    \vspace{-0.2cm}
    \begin{center}
        $\downarrow$ \textit{SCOPE Guideline Synthesis (async)} $\downarrow$
    \end{center}
    \vspace{-0.2cm}
    \begin{tcolorbox}[width=0.75\linewidth, colback=myblue!5, colframe=myblue, title=\textbf{Guideline Synthesized [Conf: 0.95]}, boxrule=1pt, arc=2pt, left=2pt, right=2pt, top=2pt, bottom=2pt]
        ``Always list all \underline{plausible label synonyms and phrase} \underline{variants} when planning extraction from figures or comparing articles.''
    \end{tcolorbox}
    \vspace{-0.2cm}
    \begin{center}
        $\downarrow$ \textit{Prompt Updated} $\downarrow$
    \end{center}
    \vspace{-0.2cm}
    \begin{tcolorbox}[width=0.75\linewidth, colback=mycoral!5, colframe=mycoral, title=\textbf{Step 4: Language Adoption}, boxrule=1pt, arc=2pt, left=2pt, right=2pt, top=2pt, bottom=2pt]
        \texttt{[19:17:54] Planning agent updates plan:}\\
        \textit{``List all \underline{plausible label synonyms and phrase variants} for the extracted axis label words...''}
    \end{tcolorbox}
    \caption{Compliance trace demonstrating \textit{language adoption}. After SCOPE synthesizes a guideline (\textcolor{myblue}{blue}), the agent's output (\textcolor{mycoral}{coral}) incorporates the exact phrasing (underlined).}
    \label{fig:compliance_trace}
\end{figure}

\begin{figure}[h]
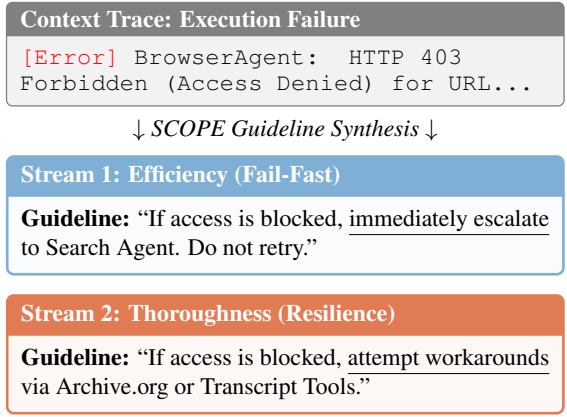

    \centering
    \small
    \begin{tcolorbox}[width=0.75\linewidth, colback=gray!10, colframe=black!50, title=\textbf{Context Trace: Execution Failure}, boxrule=0.5pt, arc=2pt, left=2pt, right=2pt, top=2pt, bottom=2pt]
        \texttt{\textcolor{red}{[Error]} BrowserAgent: HTTP 403 Forbidden (Access Denied) for URL...}
    \end{tcolorbox}
    \vspace{-0.2cm}
    \begin{center}
        $\downarrow$ \textit{SCOPE Guideline Synthesis} $\downarrow$
    \end{center}
    \vspace{-0.2cm}
    \begin{tcolorbox}[width=0.75\linewidth, colback=myblue!5, colframe=myblue, title=\textbf{Stream 1: Efficiency (Fail-Fast)}, boxrule=1pt, arc=2pt, left=2pt, right=2pt, top=2pt, bottom=2pt]
        \textbf{Guideline:} ``If access is blocked, \underline{immediately escalate} to Search Agent. Do not retry.''
    \end{tcolorbox}
    \vspace{-0.1cm}
    \begin{tcolorbox}[width=0.75\linewidth, colback=mycoral!5, colframe=mycoral, title=\textbf{Stream 2: Thoroughness (Resilience)}, boxrule=1pt, arc=2pt, left=2pt, right=2pt, top=2pt, bottom=2pt]
        \textbf{Guideline:} ``If access is blocked, \underline{attempt workarounds} via Archive.org or Transcript Tools.''
    \end{tcolorbox}
    \caption{Divergence in evolved guidelines. From the same error, Efficiency (\textcolor{myblue}{blue}) optimizes for speed, while Thoroughness (\textcolor{mycoral}{coral}) optimizes for success.}
    \label{fig:qualitative_example}
\end{figure}

This appendix provides a detailed qualitative comparison of the guidelines evolved by the Efficiency and Thoroughness optimization streams. Figure~\ref{fig:compliance_trace} demonstrates guideline compliance through language adoption, while Figure~\ref{fig:qualitative_example} illustrates the divergence between the two perspective streams.

\subsection{Strategy Patterns by Scenario}

Table~\ref{tab:qualitative_comparison} summarizes the distinct strategies that emerge from each perspective stream across common failure scenarios.

\begin{table}[h]
    \centering
    \caption{Qualitative comparison of guidelines evolved by the two perspective streams. The Efficiency stream focuses on latency reduction and fast failure recovery, while the Thoroughness stream prioritizes exhaustiveness and resilience.}
    \label{tab:qualitative_comparison}
    \resizebox{0.9\linewidth}{!}{
    \begin{tabular}{lll}
        \toprule
        \textbf{Scenario} & \textbf{Efficiency Stream Strategy} & \textbf{Thoroughness Stream Strategy} \\
        \midrule
        \textbf{Blocked Access} & Switch Agent (Escalate) & Find Alternate Source (Archive.org) \\
        \textbf{Tool Failure} & Switch Tool (Escalate) & Fix Input / Manual Fallback \\
        \textbf{Planning} & Limit Steps ($\le$ 5) & Exhaustive Case Enumeration \\
        \textbf{Goal} & Minimize Latency & Maximize Success Rate \& Auditability \\
        \bottomrule
    \end{tabular}
    }
\end{table}

\minisection{Blocked Access (HTTP 403/404)}
When the browser agent encounters access denials, the two streams diverge significantly:
\begin{itemize}[leftmargin=*,itemsep=0pt]
    \item \textbf{Efficiency}: Synthesizes guidelines to ``fail-over fast'' by immediately escalating to the web search agent, avoiding wasted retries on blocked resources.
    \item \textbf{Thoroughness}: Synthesizes guidelines to ``find alternate sources'' such as Archive.org, Wayback Machine, or transcript extraction tools, prioritizing information recovery over speed.
\end{itemize}

\minisection{Tool Failures}
For general tool execution errors:
\begin{itemize}[leftmargin=*,itemsep=0pt]
    \item \textbf{Efficiency}: Favors immediate tool switching---if one tool fails, escalate to a different tool or agent rather than debugging.
    \item \textbf{Thoroughness}: Attempts input repair or manual fallback strategies, such as reformatting queries or trying alternative parameters.
\end{itemize}

\minisection{Planning Strategy}
The streams also differ in their approach to task planning:
\begin{itemize}[leftmargin=*,itemsep=0pt]
    \item \textbf{Efficiency}: Generates guidelines limiting plan complexity (e.g., ``Keep plans under 5 steps''), favoring concise execution paths.
    \item \textbf{Thoroughness}: Generates guidelines encouraging exhaustive case enumeration and comprehensive validation before proceeding.
\end{itemize}

\subsection{Complementary Strengths}

This duality is why the Global Ensemble outperforms either individual stream. Efficiency excels at time-constrained tasks and long-horizon scenarios (Level 3) where context bloat is dangerous, while Thoroughness excels at tasks requiring deep retrieval and careful validation (Level 2). By maintaining both streams and selecting the best outcome, SCOPE adapts to heterogeneous task requirements without manual configuration.

\section{Model Choice for Meta-Agents: Detailed Analysis}
\label{app:model_ablation}

This appendix provides a detailed analysis of how different model choices for SCOPE's meta-agents (Generator, Classifier, Optimizer) affect the characteristics of evolved guidelines. We compare three configurations on the GAIA benchmark: (1) all meta-agents use GPT-4.1, (2) all use Gemini-2.5-Pro, and (3) each meta-agent uses the same model as the base agent it optimizes.

\subsection{Guideline Generation Statistics}

Table~\ref{tab:model_rule_stats} provides comprehensive statistics on the guidelines generated by each configuration.

\begin{table}[h]
    \centering
    \caption{Detailed statistics of guidelines generated by different meta-agent model configurations.}
    \label{tab:model_rule_stats}
    \resizebox{0.8\linewidth}{!}{
    \begin{tabular}{lccc}
        \toprule
        \textbf{Metric} & \textbf{GPT-4.1} & \textbf{Gemini-2.5-Pro} & \textbf{Same Model} \\
        \midrule
        Total Guidelines & 111 & 163 & 108 \\
        Avg. Guideline Length (chars) & 380.1 & 426.3 & 396.8 \\
        Max Guideline Length (chars) & 2,895 & 2,282 & 2,340 \\
        Min Guideline Length (chars) & 105 & 82 & 94 \\
        Avg. Confidence & 0.939 & 0.940 & 0.942 \\
        \midrule
        Accuracy (\%) & 46.06 & 46.67 & 45.45 \\
        \bottomrule
    \end{tabular}
    }
\end{table}

A key finding is that \textbf{Gemini generates 46\% more guidelines than GPT-4.1}, yet achieves only marginally better performance (+0.61\%). This strongly suggests that SCOPE's effectiveness stems from guideline quality rather than quantity---the Best-of-N selection and memory optimization mechanisms effectively filter for useful guidelines regardless of how many candidates are initially generated.

\subsection{Category Distribution Analysis}

Table~\ref{tab:model_category_dist} shows the distribution of guidelines across semantic categories for each configuration.

\begin{table}[h]
    \centering
    \caption{Guideline distribution by category. Each model exhibits distinct ``personality'' in guideline generation patterns.}
    \label{tab:model_category_dist}
    \resizebox{0.7\linewidth}{!}{
    \begin{tabular}{lccc}
        \toprule
        \textbf{Category} & \textbf{GPT-4.1} & \textbf{Gemini-2.5-Pro} & \textbf{Same Model} \\
        \midrule
        Analysis Methodology & 31 & 37 & 33 \\
        Data Validation & 23 & 24 & 25 \\
        Tool Usage & 28 & 36 & 26 \\
        Efficiency & 18 & 32 & 20 \\
        Error Handling & 11 & 26 & 4 \\
        General & 0 & 6 & 0 \\
        Safety & 0 & 2 & 0 \\
        \midrule
        \textbf{Total} & 111 & 163 & 108 \\
        \bottomrule
    \end{tabular}
    }
\end{table}

Gemini generates the most guidelines across nearly all categories, with notably more \textit{error\_handling} guidelines (26 vs. 4--11 for others) and \textit{efficiency} guidelines (32 vs. 18--20). It is also the only model to generate \textit{safety}-related guidelines (2). This suggests Gemini's synthesis tends toward comprehensive, verbose guidelines covering edge cases and operational robustness.

\subsection{Confidence Score Distribution}

All three configurations achieve remarkably similar average confidence scores:
\begin{itemize}[leftmargin=*,itemsep=0pt]
    \item GPT-4.1: 0.939
    \item Gemini-2.5-Pro: 0.940
    \item Same Model: 0.942
\end{itemize}

\noindent This consistency ($\sim$0.94 across all settings) suggests that all models are capable of generating guidelines they assess as reliable, regardless of their different generation patterns.

\subsection{Implications for Deployment}

The near-identical performance across configurations (within 1.2\%) has important practical implications:

\begin{enumerate}[leftmargin=*,itemsep=2pt]
    \item \textbf{Model-Agnostic Framework}: SCOPE's effectiveness is not tied to a specific meta-agent model, simplifying deployment across different infrastructure constraints.
    \item \textbf{Cost-Performance Trade-off}: Practitioners can choose meta-agent models based on cost or latency rather than accuracy. For example, using a cheaper model for the Generator may reduce costs without impacting performance.
    \item \textbf{Quality Over Quantity}: The 46\% guideline generation difference between Gemini and GPT-4.1 without proportional accuracy gains validates SCOPE's guideline management mechanisms---Best-of-N selection and memory optimization effectively filter for high-quality guidelines.
    \item \textbf{Consistent Guideline Quality}: The similar average confidence scores ($\sim$0.94) across all configurations suggest that both models are capable of generating guidelines they assess as reliable, regardless of their different generation patterns.
\end{enumerate}

\section{Additional Experiments}
\label{app:additional_exp}

\subsection{Single-Agent Evaluation: SWE-bench Verified}
\label{app:swe_bench}

To verify that SCOPE generalizes beyond multi-agent systems, we adapted SCOPE to mini-swe-agent v2~\citep{yang2024sweagent}, a single-agent software engineering system, and evaluated on SWE-bench Verified~\citep{jimenez2023swe} with GPT-5-mini. The key requirement for SCOPE is a natural step boundary (\textit{e.g.}, tool call completion), which virtually all agentic systems provide.

\begin{table}[h]
    \centering
    \caption{Single-agent evaluation on SWE-bench Verified. SCOPE provides consistent gains (+8.2 points) in a coding setup with a completely different agent architecture.}
    \label{tab:swe_bench}
    \resizebox{0.4\linewidth}{!}{
    \begin{tabular}{lc}
        \toprule
        \textbf{Method} & \textbf{SWE-bench Verified} \\
        \midrule
        Baseline & 54.2 \\
        Baseline + Reflection & 54.8 \\
        \textbf{SCOPE} & \textbf{62.4} \\
        \bottomrule
    \end{tabular}
    }
\end{table}

SCOPE provides a +8.2 point improvement, confirming that it is a framework-agnostic prompt evolution layer. The Baseline+Reflection ablation (which adds per-step LLM reflection without guideline evolution) gains only +0.6, showing that extra compute alone does not reproduce SCOPE's gains.

\subsection{Scaling to Stronger Models: GPT-5.4}
\label{app:gpt54}

A natural question is whether the failure modes SCOPE targets persist in stronger frontier models. We evaluate SCOPE with GPT-5.4 on GAIA to test this.

\begin{table}[h]
    \centering
    \caption{SCOPE with GPT-5.4 on GAIA. Despite being substantially stronger, GPT-5.4 still accumulates 1,083 errors; SCOPE reduces them by 56\% and improves accuracy by +17.5 points.}
    \label{tab:gpt54}
    \resizebox{0.3\linewidth}{!}{
    \begin{tabular}{lcc}
        \toprule
        \textbf{Method} & \textbf{GAIA} & \textbf{Errors} \\
        \midrule
        GPT-4.1 & 31.35 & 1,714 \\
        GPT-4.1 + SCOPE & 46.06 & 1,109 \\
        \midrule
        GPT-5.4 & 49.80 & 1,083 \\
        GPT-5.4 + SCOPE & \textbf{67.30} & \textbf{473} \\
        \bottomrule
    \end{tabular}
    }
\end{table}

GPT-5.4 is substantially stronger than GPT-4.1 (49.80 vs. 31.35), yet it still accumulates 1,083 errors across 165 tasks. SCOPE reduces these errors by 56\% and improves accuracy by +17.5 points. The consistent relative gain across both models confirms that the failure modes are structural limitations of static prompts, not artifacts of weaker models.

\subsection{Hyperparameter Sensitivity: N and K}
\label{app:nk_sensitivity}

We study the sensitivity of the Best-of-N candidate count ($N$) and the number of parallel perspective streams ($K$) on GAIA.

\begin{table}[h]
\centering
\begin{minipage}[t]{0.48\linewidth}
    \centering
    \caption{Sensitivity to $N$ (Best-of-N candidates) with $K=2$ fixed on GAIA.}
    \label{tab:n_sensitivity}
    \resizebox{0.8\linewidth}{!}{
    \begin{tabular}{lccc}
        \toprule
        $N$ & 1 & 2 & 3 \\
        \midrule
        GAIA Acc. (\%) & 53.38 & 56.97 & 57.85 \\
        $\Delta$ & --- & +3.59 & +0.88 \\
        \bottomrule
    \end{tabular}
    }
\end{minipage}
\hfill
\begin{minipage}[t]{0.48\linewidth}
    \centering
    \caption{Sensitivity to $K$ (perspective streams) with $N=2$ fixed on GAIA.}
    \label{tab:k_sensitivity}
    \resizebox{0.8\linewidth}{!}{
    \begin{tabular}{lcc}
        \toprule
        $K$ & 1 & 2 \\
        \midrule
        GAIA Acc. (\%) & 46.06 & 56.97 \\
        $\Delta$ & --- & +10.91 \\
        \bottomrule
    \end{tabular}
    }
\end{minipage}
\end{table}

$N=2$ captures most of the benefit over $N=1$ (+3.59\%); $N=3$ adds only +0.88\% at 50\% extra compute, justifying our choice of $N=2$. For $K$, the gain from $K=1$ to $K=2$ is +10.91\%, reflecting strategy diversity rather than an extra selection opportunity: all methods use Pass@2 (2 independent runs per task), so baselines run the same static prompt twice while SCOPE runs 2 differently evolved prompts (Efficiency vs. Thoroughness) once each.

\subsection{Step Budget Ablation}
\label{app:step_budget}

A potential confound is that SCOPE's gains come from agents having insufficient steps. We test this by giving the baseline 50\% more steps.

\begin{table}[h]
    \centering
    \caption{Impact of increasing step budget. 50\% more steps yields negligible improvement, confirming the root cause is systematic errors from static prompts, not insufficient budget.}
    \label{tab:step_budget}
    \resizebox{0.7\linewidth}{!}{
    \begin{tabular}{lccc}
        \toprule
        \textbf{Method} & \textbf{GAIA} & \textbf{DeepSearch} & \textbf{Step-limit hits} \\
        \midrule
        Baseline & 32.73 & 14.00 & 255 \\
        Baseline (50\% more steps) & 33.13 & 14.00 & 250 \\
        \textbf{SCOPE} & \textbf{56.97} & \textbf{32.00} & \textbf{139} \\
        \bottomrule
    \end{tabular}
    }
\end{table}

50\% more steps yields only +0.40\% on GAIA and zero on DeepSearch. Step-limit hits barely change (255 to 250), showing the agent repeats the same errors regardless of budget. SCOPE cuts step-limit hits by 45\% while achieving better performance with fewer steps.

\subsection{Classifier Threshold Sensitivity}
\label{app:threshold}

The confidence threshold $c_{\text{thresh}}$ controls whether a guideline is promoted to strategic memory (persistent) or kept as tactical (task-specific). We validate the choice of 0.85 through a sensitivity experiment on GAIA.

\begin{table}[h]
    \centering
    \caption{Classifier threshold sensitivity on GAIA. A clear peak at 0.85 indicates meaningful routing decisions.}
    \label{tab:threshold}
    \resizebox{0.7\linewidth}{!}{
    \begin{tabular}{lccccc}
        \toprule
        \textbf{Threshold} & 0.50 & 0.75 & \textbf{0.85} & 0.90 & 0.95 \\
        \midrule
        GAIA Acc. (\%) & 47.87 & 53.93 & \textbf{56.97} & 51.51 & 46.66 \\
        \bottomrule
    \end{tabular}
    }
\end{table}

The clear peak at 0.85 with degradation at both extremes validates the classifier: too low (0.50) promotes noisy guidelines to strategic memory, while too high (0.95) prevents useful guidelines from persisting. In practice, the classifier is highly selective: only 320 of 1,570 synthesis attempts (20.4\%) are promoted to strategic memory, with 79.2\% of rejections being duplicate detection.

\end{document}